\documentclass{article}
\pdfoutput=1

\usepackage[square,numbers]{natbib}
\usepackage[nonatbib,preprint]{neurips_2023}

\usepackage[utf8]{inputenc} % allow utf-8 input
\usepackage[T1]{fontenc}    % use 8-bit T1 fonts
\usepackage{hyperref}       % hyperlinks
\usepackage{url}            % simple URL typesetting
\usepackage{booktabs}       % professional-quality tables
\usepackage{amsfonts}       % blackboard math symbols
\usepackage{nicefrac}       % compact symbols for 1/2, etc.
\usepackage{microtype}      % microtypography
\usepackage{xcolor}         % colors
\usepackage{tabularx}
\usepackage{subfig}
\usepackage{graphicx}
\usepackage{multirow}
\usepackage{wrapfig}

\title{Leaving Reality to Imagination: Robust Classification via Generated Datasets}
\author{
    Hritik Bansal\\ UCLA \\ \texttt{hbansal@ucla.edu}
    \And
    Aditya Grover \\ UCLA \\ \texttt{adityag@cs.ucla.edu}
}

\begin{document}

\maketitle

\begin{abstract}
Recent research on robustness has revealed significant performance gaps between neural image classifiers trained on datasets that are similar to the test set, and those that are from a naturally \textit{shifted} distribution, such as sketches, paintings, and animations of the object categories observed during training. Prior work focuses on reducing this gap by designing engineered augmentations of training data or through unsupervised pretraining of a single large model on massive in-the-wild training datasets scraped from the Internet. However, the notion of a dataset is also undergoing a paradigm shift in recent years.
With drastic improvements in the quality, ease-of-use, and access to modern generative models,  generated data is pervading the web. In this light, we study the question: How do these generated datasets influence the natural robustness of image classifiers? We find that Imagenet classifiers trained on real data augmented with generated data achieve higher accuracy and effective robustness than standard training and popular augmentation strategies in the presence of natural distribution shifts. We analyze various factors influencing these results, including the choice of conditioning strategies and the amount of generated data. Additionally, we find that the standard ImageNet classifiers suffer a performance degradation of upto $20\%$ on the generated data, indicating their fragility at accurately classifying the objects under novel variations. Lastly, we demonstrate that the image classifiers, which have been trained on real data augmented with generated data from the base generative model, exhibit greater resilience to natural distribution shifts compared to the classifiers trained on real data augmented with generated data from the finetuned generative model on the real data. The code, models, and datasets are available at \url{https://github.com/Hritikbansal/generative-robustness}.

% we benchmark the performance of the standard ImageNet classifiers on the generated data, and find that there is a significant gap between their performance 

% Lastly, we introduce and analyze an evolving generated dataset, ImageNet-G-v1, to better benchmark the design, utility, and critique of standalone generated datasets for robust and trustworthy machine learning.
\end{abstract}
\section{Introduction}
\label{introduction}

The ultimate goal of machine learning is to create models that can generalize beyond their training data. However, recent studies \cite{recht2019imagenet,hendrycks2021many,wang2019learning,barbu2019objectnet,taori2020measuring} have shown a gap between the performance of deep neural classifiers on test data that is independent and identically distributed (i.i.d.) as the training data, and \textit{shifted} datasets containing natural variations of the images in the training distribution. For instance, a ResNet-101 \cite{he2016deep} model trained on ImageNet-1K \cite{deng2009imagenet} experiences a $50\%$ reduction in the performance when evaluated on ImageNet-Sketch \cite{wang2019learning}, a dataset of sketches of objects from ImageNet classes. This fragility of classifiers limits their use in real-world applications such as autonomous driving and medical diagnosis.

% \ag{potentially better flow: increasing evidence that availability of larger datasets can mitigate this brittleness. 2 ways of doing it: data augs, get more in-the-wild diverse data. modern generative models can get the best of both worlds. they are trained on large, diverse datasets and they are not restricted to fixed, finite set of hand engineered augmentations --- a. stochastic models queried as many times as you want + b. flexible in how you draw these generated augmentations (using text prompts, images, with / without guidance etc. + c. zero-shot generalize to arbitrary regimes) }

One effective strategy to improve robustness is to enlarge the amount of training data by designing intricate augmentations~\cite{hendrycks2019augmix,hendrycks2022pixmix,hendrycks2021many} of the training data that aid the generalization of classifier to novel domains. 
% However, discovering and implementing these augmentation strategies can be challenging and may rely on unrealistic or hard-to-understand perturbations. 
Similarly, datasets can also be enlarged by scraping multimodal paired datasets, such as image-caption pairs on the Internet~\cite{radford2021learning,jia2021scaling,pham2021combined}. However, the notion of a dataset is also experiencing a paradigm shift in recent years. With the emergence of modern `in the wild' generative models \cite{ramesh2022hierarchical,nichol2021glide,rombach2022high,saharia2022photorealistic,chang2023muse},
generated data is pervading the web \cite{wang2022diffusiondb,kirstain2023pick}. These models are trained on large diverse datasets \cite{schuhmann2022laion} with open vocabulary annotations, such that post-training, they can synthesize high-fidelity images for a wide range of concepts in a \textit{zero-shot} manner. Notably, these models are not limited to generate a fixed, finite set of hand-engineered augmentations and can be repeatedly queried to generate diverse data through various conditioning mechanisms such as text prompts, images, and guidance strategies. 

\begin{figure}[h]
    \centering
    \includegraphics[width=0.7\linewidth]{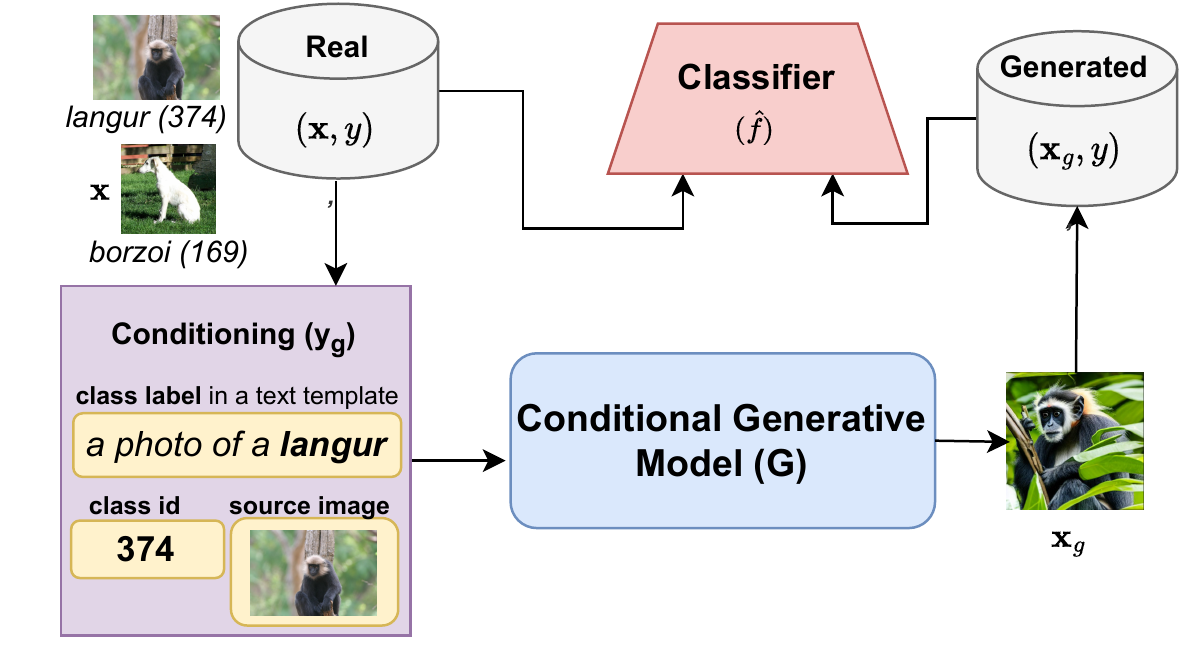}
    \caption{Overview of our approach. Our method creates generated dataset using a conditional generative model. The real dataset is then augmented with the generated dataset to train a classifier.}
    \label{fig:data_gen}
\end{figure}

 % However, 
 In this work, we study the question: How do datasets generated from modern in-the-wild generative models influence the natural robustness of image classifiers? 
 Specifically, we focus on the classification accuracy \cite{ravuri2019classification}, and the effective robustness \cite{taori2020measuring} of the standard classifiers trained from scratch. We present an overview of our setup in Figure \ref{fig:data_gen}.
For generating data, we utilize Stable Diffusion \cite{rombach2022high}, an in-the-wild, open-source conditional generative model and create a synthetic dataset conditioned on objects from two source datasets ImageNet-1K \cite{deng2009imagenet} and ImageNet-100 \cite{tian2020contrastive}. By repeatedly sampling from Stable Diffusion by prompting it with diverse captions for the class labels, we generate  a large and diverse synthetic dataset. Specifically, we generate 1.3M synthetic images for training and 50K images for validation, which is the same size as the real ImageNet-1K training and validation data.
 This complements concurrent works on using synthetic data for augmentating and improving the accuracy of contrastive methods~\cite{he2022synthetic,radford2021learning} on image classification and
% In our work, we focus on the benefits of the generated data acquired zero-shot from the generative model without any finetuning of the base generative model using the real images, whereas there are
other works \cite{trabucco2023effective,azizi2023synthetic} that study generative augmentations post-finetuning of the part or whole of the generative model on the real data distribution. 
Our work focusses on the more challenging setting of transfer to image classifiers without any finetuning of the base generative model on the real images. We provide further comparison with the change in the data generation paradigm in \S \ref{sd-finetune-im1k}.

Our main takeaway is that training a classifier on a combination of real and generated data can achieve high absolute performance and high effective robustness (\S \ref{exp:robustness}) on natural distribution shift datasets. Removing either real or generated data results in a corresponding reduction in accuracy and effective robustness respectively, thus necessitating the use of a mixture. Previous work \cite{yuan2022not} shows that we can manipulate the generative models to adapt the images from a source domain to a single target domain which results in accurate classifiers on the target domain. However, in our work, we create a single generated dataset from a diverse set of templates without customizing it to a single target domain. 

To further explain our results, we find that the `in-the-wild' aspects of modern generative indeed plays a role and substituting these generations with hand-crafted augmentation strategies or outputs of traditional class-conditional generative models is less effective (\S \ref{exp:benchmark}). We supplement this analysis with additional results on the impact of proportion sizes of real and generated data (\S \ref{exp:training_size}), different multimodal conditioning strategies for data generation (\S \ref{exp:gen_strategy}), and a human and automatic evaluation study to assess and compare the class consistency, image quality, and diversity of the real and generated images (Appendix \S \ref{appen:human_eval}). Having studied the utility of the generated datasets for training, we study their use case for benchmarking the standard ImageNet classifiers. In \S \ref{exp:imnet_g_eval}, we find that the classifiers such as ResNet-101 \cite{he2016deep}, finetuned CLIP \cite{radford2021learning,wortsman2022robust} and Vision Transformers \cite{dosovitskiy2020image,tu2022maxvit} suffer an absolute degradation of $20\%$ on the generated data created using text prompts with the class labels, suggesting their fragility to newly generated natural variations. 

Finally, we study the impact of varying the data generation paradigm and evaluate the quality of the image classifiers trained on the generated data that is closer in distribution to the real data as compared to the generated data collected in a zero-shot way. In \S \ref{sd-finetune-im1k}, we find that training the image classifier on the real data augmented with the generated data from the base generative model achieves high accuracy on the natural distribution shift datasets than training it on the real data augmented with the generated data synthesized from the finetuned generative model on the real ImageNet data. Our base generated and finetuned generated datasets are made publicly available allowing for easy and reproducible benchmarking of utility and critique of the generated datasets.

\section{Background}
\label{background}

\subsection{Supervised Classification}

Given a labelled dataset $\mathcal{D} = \{(\mathbf{x}_1, y_1), \ldots ,(\mathbf{x}_n, y_n)\} \sim P(\mathbf{x}, y)$ where $\mathbf{x}_{i} \in \mathcal{X} \subset \mathcal{R}^{d}$ represents the $i^{th}$ input, and $y_i \in \mathcal{Y} \subset \{1,2,\ldots,\mathcal{K}\}$ represents its corresponding target label, we train a classifier $\hat{f}(\mathbf{x})$ on $\mathcal{D}_{train} \subset \mathcal{D}$ such that it models  $P(y|\mathbf{x})$, i.e., conditional distribution of $y$ given the input $\mathbf{x}$. The classification model is usually trained via empirical risk minimization, $L(\hat{f}, \mathcal{D}_{train}) = \mathbb{E}_{(x,y)\sim D_{train}}\left[l(\hat{f}(\mathbf{x}), y)\right]$, where $l$ is the training objective, under the assumption that samples in the training data are identically and independently distributed (i.i.d.). Eventually, we evaluate the performance of the classifier on a held test set $D_{test} \subset \mathcal{D} \sim P$ with ($\mathcal{D}_{test}\cap \mathcal{D}_{train}=\phi$) using \textit{accuracy} $A(\hat{f}, \mathcal{D}_{test}) = \mathbb{E}_{(x,y) \sim D_{test}}\left[\mathbb{I}(\hat{f}(\mathbf{x}) = y)\right]$. 

If a classifier achieves high accuracy on the examples from the test set, we hope that it will perform well on the other examples that come from $P$ as well as semantically related data distributions. However, in practice, we encounter test sets $\mathcal{D}'$ sampled from a data distribution $P'$ that contains the samples resembling the ones in $\mathcal{D}$ with slight variations e.g., images in $\mathcal{D}'$ may vary from the images in the $\mathcal{D}$ in terms of differences in camera settings, and captured views.

\subsection{Robustness}

For any classifier, we can quantify the \textit{accuracy gap} (AG) between the accuracy on a test set that follows the same distribution as the training set, and a test set that varies naturally from the training distribution. 

\begin{equation}
\label{eq:accuracygap}
    AG(\hat{f}, \mathcal{D}_{test}, \mathcal{D}') =   A(\hat{f}, \mathcal{D}') - A(\hat{f}, \mathcal{D}_{test})
\end{equation}

For a robust classifier, the accuracy gap should be low up to random sampling error. However, a classifier that closes the accuracy gap might decrease the individual accuracies. Additionally, given a robust classifier $\hat{f}$ that offers high accuracy on the shifted datasets, we can assess it relative to the expected accuracy on the shifted dataset with a standard classifier that is trained on the source training set without any specific robustness intervention. This notion is formalized as \textit{effective robustness} (ER)~\citep{recht2019imagenet,recht2018cifar}.

\begin{equation}
\label{eq:effectiverobustness}
    ER(\hat{f},\mathcal{D'}, \mathcal{D}_{test}) = A(\hat{f}, \mathcal{D}') - \beta(A(\hat{f}, \mathcal{D}_{test}), \mathcal{D}', \mathcal{D}_{test})
\end{equation}
    
where $\beta(z, \mathcal{D}', \mathcal{D}_{test})$ is the accuracy on the shifted test set $\mathcal{D}'$ for a given accuracy $z = A(\hat{f}, D_{test})$ on the source test set $\mathcal{D}_{test}$. We calculate $\beta$ by fitting a linear function on the collection of standard classifiers. Positive ER indicates that the robustness intervention improves over standard training.
% Hence, a robustness intervention on classifiers should improve its effective robustness over standard training.

\subsection{Generative Modeling}

% To build robust classifiers, several studies have argued for the effectiveness of large training datasets that have been achieved by either highly-engineered data augmentations \cite{hendrycks2019augmix,cubuk2018autoaugment}, or diverse in-the-wild pretraining paradigms \cite{radford2021learning,jia2021scaling,pham2021combined}. To this end, we can leverage the stochasticity, flexibility of design, and zero-shot capabilities of the modern generative models \cite{ramesh2022hierarchical,nichol2021glide,rombach2022high} and many others to create a generated diverse in-the-wild dataset. 

Generative models $p_{\theta}(\mathbf{x})$ are probabilistic models that are trained to learn the data distribution $p_{data}(\mathbf{x})$ \cite{tomczak2022deep}. Due to their flexible design, we can further train their class-conditional versions \cite{brock2018large,karras2019style} to model the class-conditional distributions $p(\mathbf{x}|y_g)$ where is $y_{g}$ is the conditioning variable, that can take various forms, which we describe in next section. Post-training, we can generate a new sample $\mathbf{x}_g$ by sampling from the class-conditional model distribution $\mathbf{x}_g \sim p_{\theta}(\mathbf{x}|y_g)$. In Figure 1, this stochastic mapping $p_{\theta}(\mathbf{x}|y_g)$ is referred to as $G$. 

%Thus, we can create a generated dataset $\mathcal{D}_{g} = \{(\mathbf{x}_g, y)\}$ by repeatedly querying the conditional generative model.%\ag{class-conditional is typically exclusive term for conditioning on class-idx. just say conditional} 

% \ag{vector notation is inconsistent here and even following sections. make x's bold everywhere (or nowhere if you want to not bother with emphasizing vectors at all).}
% \ag{for y's, you will have to introduce different symbol when used as conditioning vs when used as label for classifier. you can say perhaps $y_g$ or something and say there is some mechanism for relating $y_g$ to $y$ with examples (which we describe in next section). make sure to fix Figure 1 as well after these changes. }

\subsection{Data Generation using Stable Diffusion}
\label{background:sd}

% In this work, we employ Stable Diffusion (SD) \cite{rombach2022high}, an `in the wild' generative model is one that can generate images from the natural language description of a wide range of concepts, combine unrelated concepts in a realistic manner, and apply novel transformations to existing images. Such abilities are exhibited by Stable Diffusion through training on a large, diverse dataset LAION \cite{schuhmann2022laion} on matched image-text pairs scraped from the web. 

Given a single data point $(\mathbf{x}, y)$ from the source dataset, we have various ways to generate a new data point $\mathbf{x}_g$ with a trained Stable Diffusion \cite{rombach2022high}, as summarized in Appendix Figure \ref{appen_fig:gen_strategies}.

\textbf{Generation via Class Labels}: Here, we synthesize images by conditioning on the natural language templates $\mathcal{M}$ for the class labels $y$. An example template $M(y)$ = `a photo of a $y$' where $M \subset \mathcal{M}$ and $y$ is the class label. Hence, the proxy caption for a `dog' class label with the template $M$ would be `A photo of a \textit{dog}'. This generation strategy involves using a pretrained CLIP text encoder $y_g = CLIP_{text}(M(y))$. Since generating data conditioned on the natural text descriptions is the default setting for data generation using Stable Diffusion, our primary focus is on the natural robustness elicited by this data generation strategy. In addition to the traditional zero-shot data generation approach, we study the following other ways to generate images without any training or finetuning of the generative model on the images from the source dataset. We specifically study the effect of these data generation procedures in \S \ref{exp:gen_strategy}. 

\textbf{Generation via Real (Source) Images:} Since CLIP text and image embeddings are aligned in the representation, in principle, they can be used interchangeably. Here, we use CLIP's vision encoder $y_g = CLIP_{image}(\mathbf{x})$ for conditioning.\footnote{We use the implementation in \url{https://github.com/huggingface/diffusers/blob/main/src/diffusers/pipelines/stable_diffusion/pipeline_stable_diffusion_image_variation.py}} 

\textbf{Generation via Real (Source) Images and Class Labels:} We create realistic variations of the source image $\mathbf{x}$ by sampling from a noisy latent representation that is conditioned on the embedding of the source image, conditioned on the natural description of its class label $y_g = CLIP_{text}(M(y))$.

We present additional details regarding the data generation process in Appendix \S \ref{appen_data_gen_sd}. Moreover, we conduct a comprehensive quantitative comparison between the generated data and the real data, focusing on dimensions such as quality, consistency, and diversity. This evaluation is performed through both human assessment and automatic evaluation, as described in Appendix \S \ref{appen:human_eval}.

\section{Setup}
\label{setup}

\textbf{Real Dataset:} The ImageNet-1K dataset is widely used as a benchmark for building robust classifiers for image recognition. It contains 1.3 million labeled training images and 50,000 validation images across 1000 categories. To evaluate the effectiveness of generated data in this task, we use ImageNet-1K as our benchmark. However, due to the limitations of compute and storage, we also utilize ImageNet-100, a subset of 100 classes randomly sampled from ImageNet-1K, for many of our analysis and ablation studies. In line with previous studies~\cite{saha2022backdoor,tian2020contrastive}, we find that the trends observed in ImageNet-100 are similar to those in ImageNet-1K.

\textbf{Natural Distribution Shift Datasets:} Similar to the previous studies \cite{miller2021accuracy,radford2021learning,nguyen2022quality}, we consider ImageNet as the reference dataset where ImageNet-Sketch \cite{wang2019learning}, ImageNet-R \cite{hendrycks2021many}, ImageNet-V2 \cite{taori2020measuring}, and ObjectNet \cite{barbu2019objectnet} are natural distribution shift datasets. We provide more further description about these datasets in Appendix \S \ref{appen:nds}.

\textbf{Classifiers:}  We consider models with varying architectures and model capacities as classifiers. This includes ResNet-18 \cite{he2016deep}, ResNeXt-50, ResNeXt-101  \cite{xie2017aggregated}, EfficientNet-B0 \cite{tan2019efficientnet} and MobileNet-V2 \cite{howard2017mobilenets}. We provide further details on training them in Appendix \S \ref{appen:setup_train_classifier}.

\begin{figure}
    \centering
    \includegraphics[width=0.7\linewidth]{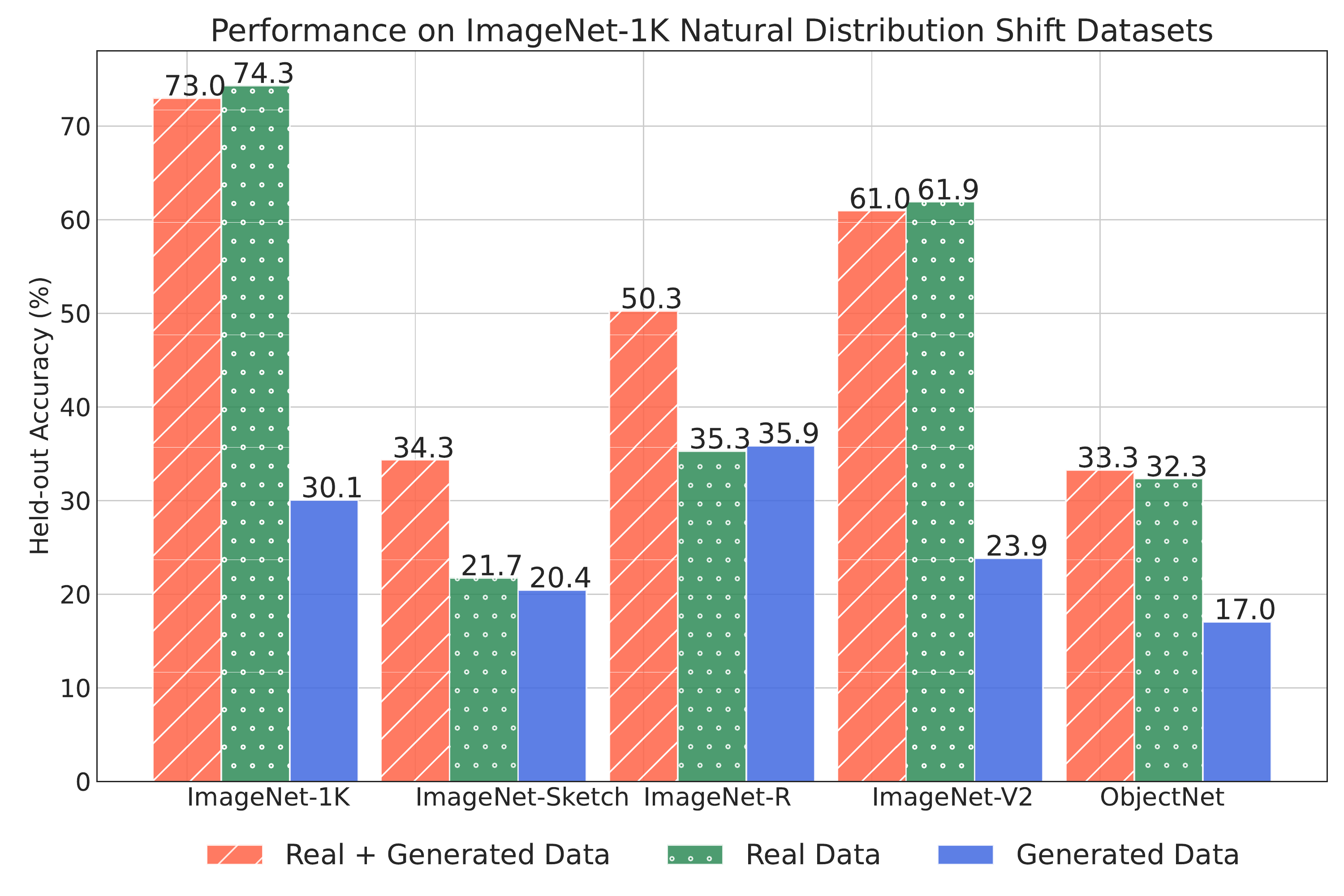}
    \caption{Accuracy of ImageNet-1K classifiers on its validation set and its natural distribution shift (NDS) datasets. The classifiers are trained (a) solely on the real data (Green), (b) solely on the generated data of equal size as the real data (Blue), and (c) full real data augmented with the complete generated data (Orange). We find that the classifiers trained with data augmentation either match or outperform the classifiers trained with just the real or generated data on various NDS datasets. The standard deviation of the accuracies ranged from $0.2$ - $1$ points for three random seeds.}
    \label{exp_fig:im1k}
\end{figure}

\textbf{Data Generation:} We utilize Stable Diffusion \cite{rombach2022high} to generate synthetic data conditioned on the natural descriptions of the objects in the dataset, and/or the training images. Specifically, we use the Stable Diffusion-V1-5 implementation and inference settings detailed in the diffusers \cite{von-platen-etal-2022-diffusers} library. For ImageNet-1K, we construct a 1.3M generated training dataset and 50K validation dataset from Stable Diffusion by conditioning on the proxy captions for the class labels. The proxy captions are a set of 80 diverse templates given by \cite{radford2021learning} to evaluate their CLIP model (Appendix Table \ref{appendix:templates}). We provide further details in Appendix \ref{appen:setup}.
% Thus, in total, we generated 390K synthetic training images and 150K validation images. 

% For SD-labels generation, we use the same set of 80 templates to create the proxy captions for the class labels. For SD-images generation, we generate a variation of the source image by sampling from the diffusion model conditioned on its CLIP embeddings. For SD-Labels\&Images, we first encode the image from the source dataset and denoise it conditioned on proxy captions constructed out of the 80 diverse templates followed by the decoder to create a new variation of the source image. 
% We further provide more details on the generation comparison in \textcolor{brown}{Appendix X/Table Y}.
\section{Experiments}
\label{experiments}

In this section, we present a comprehensive set of experiments on the usefulness of the generated data for training and evaluating robust classifiers. In \S \ref{exp:robustness}, we show that the classifiers trained with the combination of the real and generated data achieve high accuracy and effective robustness on natural distribution shift datasets. In \S \ref{exp:benchmark}, we show that the generated data, created in a zero-shot manner, is competitive or better than the modern augmentation strategies such as DeepAugment \cite{hendrycks2021many} on NDS benchmarks. In \S \ref{exp:training_size}, we show that the effective robustness is positively correlated with the size of the generated dataset. Hence, given a flexible compute budget, the practitioners should aim to sample large generated datasets for augmentations. In \S \ref{exp:gen_strategy}, we compare many flexible ways to condition the modern generative models, and show that using text conditioning with diverse set of templates (as opposed to repeated sampling from fewer templates) is best for training downstream robust classifiers. Beyond the use of the generated data for training, in \S \ref{exp:imnet_g_eval}, we benchmark a variety of ImageNet classifiers on the generated validation dataset and we find that these classifiers do not robustly classify the objects in the generated domain.

\subsection{Classification Accuracy and Robustness}
\label{exp:robustness}

We evaluate the accuracy and effective robustness of classifiers trained with different datasets on natural distribution shifts (NDS) benchmarks, including ImageNet-Sketch, ImageNet-R, ImageNet-V2, and ObjectNet. We train on 3 kinds of datasets: the \textbf{real} ImageNet-1K dataset with 1.3M images, a \textbf{generated} training dataset of 1.3M images created using Stable Diffusion conditioned on proxy captions for the class labels in ImageNet-1K, and a combination of all images from the \textbf{real and generated} training datasets. 
% As Stable Diffusion is not aware of the source data distribution, the image generation conditioned on objects' natural language descriptions is likely to synthesize diverse in-the-wild images with realistic variations that are not well-represented in the source dataset.

% \subsubsection{Training on Real and Generated Data}
% \label{exp:train_on_gen}

The average accuracy of the image classifiers over three random seeds is shown in Figure \ref{exp_fig:im1k}. We find that models trained on the real ImageNet-1K (Im-1K) dataset (Green bar) perform well on its validation set but experience a significant drop in performance under natural shifts. Interestingly, we find that training on generated images using the same training dataset size leads to poor absolute performance on Im-1K ($30\%$) as well as its NDS datasets. The low absolute performance may be due to the large distribution gap between the source and generated training datasets. However, we observe that the accuracy gaps performance on the real validation dataset and its NDS datasets are low, which might be attributed to the benefits of training on diverse generated data. Finally, we train the classifiers on an equal-sized combination of real and generated datasets to understand the effectiveness of generative augmentations.

As shown in Figure \ref{exp_fig:im1k}, we find that the absolute performance of the classifiers trained on the real data augmented with the generated data either matches or outperforms the classifiers trained solely on the real or generated dataset across all the natural distribution datasets. Notably, training on the combination of the real and generated dataset does not affect performance on the ImageNet1K validation dataset compared to standard training. We see a similar effect for the natural distribution dataset, ImageNet-V2, which is closest in distribution to ImageNet-1K since both the datasets are derived from Flickr30K \cite{recht2019imagenet}. On ObjectNet, the gain is $\sim 1\%$, indicating the difficulty of this dataset. Surprisingly, we find that training with the combination of the real and generated data leads to an absolute improvement of $\sim15\%$ on ImageNet-Sketch and ImageNet-R over standard training.

\begin{table}[h]
\begin{center}
\caption{Effective robustness of the classifiers trained on the generated dataset, and the real data augmented with the generated dataset. The results are averaged over five classifiers trained with three random seeds. The values greater than 0 indicate improvements over the standard training.}
\resizebox{\textwidth}{!}{%
\begin{tabular}{lcccc|c}\hline
           & ImageNet-Sketch & ImageNet-R & ImageNet-V2 & ObjectNet & Average \\\hline
Generated Data        & \textbf{37.8}          & \textbf{45.3}     & \textbf{9.1}       & \textbf{49.9} &  \textbf{35.6}\\
Real + Generated Data & 14.9          & 16.7     & 0.5       & 2.3  &  8.6\\\hline

\end{tabular}%
}
\label{exp_table:eff_rob_im1k}    
\end{center}
\end{table}

To benchmark the improvements against standard training, we calculate the effective robustness \cite{taori2020measuring} of the classifiers trained with the generated data. As shown in Table \ref{exp_table:eff_rob_im1k}, we find that the effective robustness of the generated data is high across all the shifted datasets (Row 1). Additionally, we find that the effective robustness (ER) of the classifier trained with the combination of the real and generated data is higher than standard training (= 0) while being lower than classifiers trained on the generated data (Row 1 and Row 2).

\begin{wraptable}{r}{9cm}
\caption{FID score averaged over the ImageNet-1K classes between the real/generated datasets and NDS datasets.}
\begin{center}
\begin{tabular}{lcccc}
\hline
              & Im-Sketch & Im-R & Im-V2 & ObjectNet \\\hline
Real Data      & 248             & 225       & \textbf{179}        & \textbf{224}     \\
Generated Data & \textbf{210}         & \textbf{190}     & 223       & 255    \\\hline
\end{tabular}
\end{center}
\label{exp:fid}
\end{wraptable}

To better understand the disparity in the gains for different NDS datasets, we calculate the Fréchet inception distance (FID) \cite{heusel2017gans} between the real (generated) images and each of the NDS datasets, averaged over each class. In Table \ref{exp:fid}, we find that the distribution gap between ImageNet-Sketch and ImageNet-R datasets is lesser to the generated data than it is to the real data. We attribute this observation to the presence of rendition and sketch images through their respective templates during the data generation process (Appendix \S \ref{appendix:templates}), which eventually gets reflected as larger improvements in classification accuracy and ER on these datasets. In \S \ref{exp:gen_strategy}, we perform experiments to better understand the effect of the data generation templates on the performance of the image classifiers. Despite the reduced distribution gap, we note that training the classifiers solely with the generated data is not enough for high accuracy on ImageNet-Sketch/Rendition (Figure \ref{exp_fig:im1k}). Hence, it implies that we do need real data in the training mix to achieve higher absolute accuracy.

In summary, generated data alone increases the effective robustness at the cost of accuracy, whereas an augmented mixture of real and generated data strikes a good balance for robust and accurate training. Even though our work is focussed on robustness to `natural' distribution shifts, our experiments show that training a classifier on the real data augmented with the generated data achieves high accuracy and ER on `synthetic' corruption-based datasets such as ImageNet-C \cite{hendrycks2019benchmarking} (Appendix \S \ref{appen:Im-C}).

\setlength{\tabcolsep}{2pt}
\begin{table}[h]
\begin{center}
\caption{Comparison of the models trained on real data and an equal mix of real data and generated data (100:100 ratio) using different augmentation strategies on ImageNet-100 validation set and its natural distribution shift (NDS) datasets. We report results over the classes that overlap with ImageNet-100. The results are averaged over three runs of ResNet-18, ResNeXt-50/101.}
\resizebox{0.8\textwidth}{!}{%
\begin{tabular}{lccccc|c}
\hline
& Im-100 & Im-Sketch-100 & Im-R-100 & Im-V2-100 & Obj-100 & Average \\ \hline
Real Data           & 85.7    & 28.4          & 49.8     & 74.8      & 42.3 & 56.2    \\
+ DeepAugment \cite{hendrycks2021many}   & 86.7    & 45.2          & 67.2     & 76.5      & 44.9 & 64.1     \\
+ PixMix \cite{hendrycks2022pixmix}   & 85.3 & 32.7 & 56.6 & 73.7 & 43.9 & 58.5     \\
+ Class Conditioned LDM \cite{rombach2022high} & 86.7    & 27.9          & 55.0     & 75.6      & 46.1 & 58.3    \\\hline
+ Stable Diffusion \cite{rombach2022high} (Ours)     & 86.8    & 48.4         & 71.2    & 76.0      & 47.5 & \textbf{66.0}    \\\hline
\end{tabular}%
}
\label{exp_table:augmentation strategies}
\end{center}
\end{table}

\setlength{\tabcolsep}{2pt}
\begin{table}[h]
\begin{center}
\caption{Comparison of the models trained on real data and an equal mix of real data and generated data (100:100 ratio) using different generation strategies on ImageNet-100 validation set and its natural distribution shift (NDS) datasets. We report results over the classes that overlap with ImageNet-100. The results are averaged over three runs. We abbreviate ImageNet as Im, and Class Label as CL.}
\resizebox{\textwidth}{!}{%
\begin{tabular}{l|ccccc|c}
\hline
&  Im-100 & Im-Sketch-100 & Im-R-100 & Im-V2-100 & Obj-100 & Average \\ \hline
Real data      &    85.7    & 28.6          & 49.8     & 74.8      & 42.3 &   56.2   \\
+ Generated data via Class labels (`a photo of a [CL]' template)  & 87.4 & 35.7 & 59.5 & 75.6 & 44.9 & 60.6\\
+ Generated data via Class labels (`a rendition of a [CL]' template)  & 87.4 & 46.3 & 67.8 & 76.0 & 46.5 & 64.8\\
+ Generated data via Class labels (80 diverse templates) &  86.8    & 48.4          & 71.2     & 76.0      & 47.5 & \textbf{66.0}   \\
+ Generated data via Real images   & 85.9 & 32.2 & 50.0 & 74.9 & 45.1 & 59.5\\
+ Generated data via Real images and Class labels & 87.4 & 46.7 & 71.4 &  76.5 & 47.9 & \textbf{66.0}\\\hline
\end{tabular}%
}
\label{exp_table:generation strategies}
\end{center}
\end{table}

\subsection{Comparison with Standard Augmentations}
\label{exp:benchmark}

How much of the above improvements can be attributed to modern `in-the-wild' generative models as opposed to traditional data augmentation paradigms? To evaluate this question, we generate new training datasets using a state-of-the-art augmentation strategy, DeepAugment \cite{hendrycks2021many}, PixMix \cite{hendrycks2022pixmix}, and a class-conditional latent diffusion model (LDM) \cite{rombach2022high} trained on ImageNet-1K alone.

We examine the average performance of three classifiers (ResNet-18, ResNeXt-50, and ResNeXt-101) trained on the real ImageNet-100 dataset with 130K images, augmented with an equal number of generated images from different generation strategies on the set of overlapping classes with 4 NDS datasets: Im-Sketch, Im-R, Im-V2, and ObjectNet in Table \ref{exp_table:augmentation strategies}. We find that the performance for all approaches are similar to standard training on the ImageNet-100 validation dataset. However, performance on NDS datasets varies greatly. We observe that augmenting with the diverse in-the-wild generated datasets yields the highest performance on ImageNet-R, ImageNet-Sketch, and ObjectNet, followed by DeepAugment. The significant difference in performance between the LDM and Stable Diffusion results across all the shifted datasets highlights the utility of modern generative models that are trained on larger multimodal datasets and allow for more flexible conditioning.
% DeepAugment slightly outperforms in-the-wild augmentation on ImageNet-V2.  The creation of a synthetic training dataset opens the door for re-evaluating the usefulness of various data augmentation strategies to train robust classifiers, which we leave as a relevant future direction of study.

% Effective robustness is only explicitly important when comparing models with lower performance on the ImageNet dataset to better compare where the trends lie. 

\begin{figure}[h]
    \centering    \subfloat[\centering\label{exp_fig:av_acc} Average Accuracy]{{\includegraphics[width=0.4\linewidth]{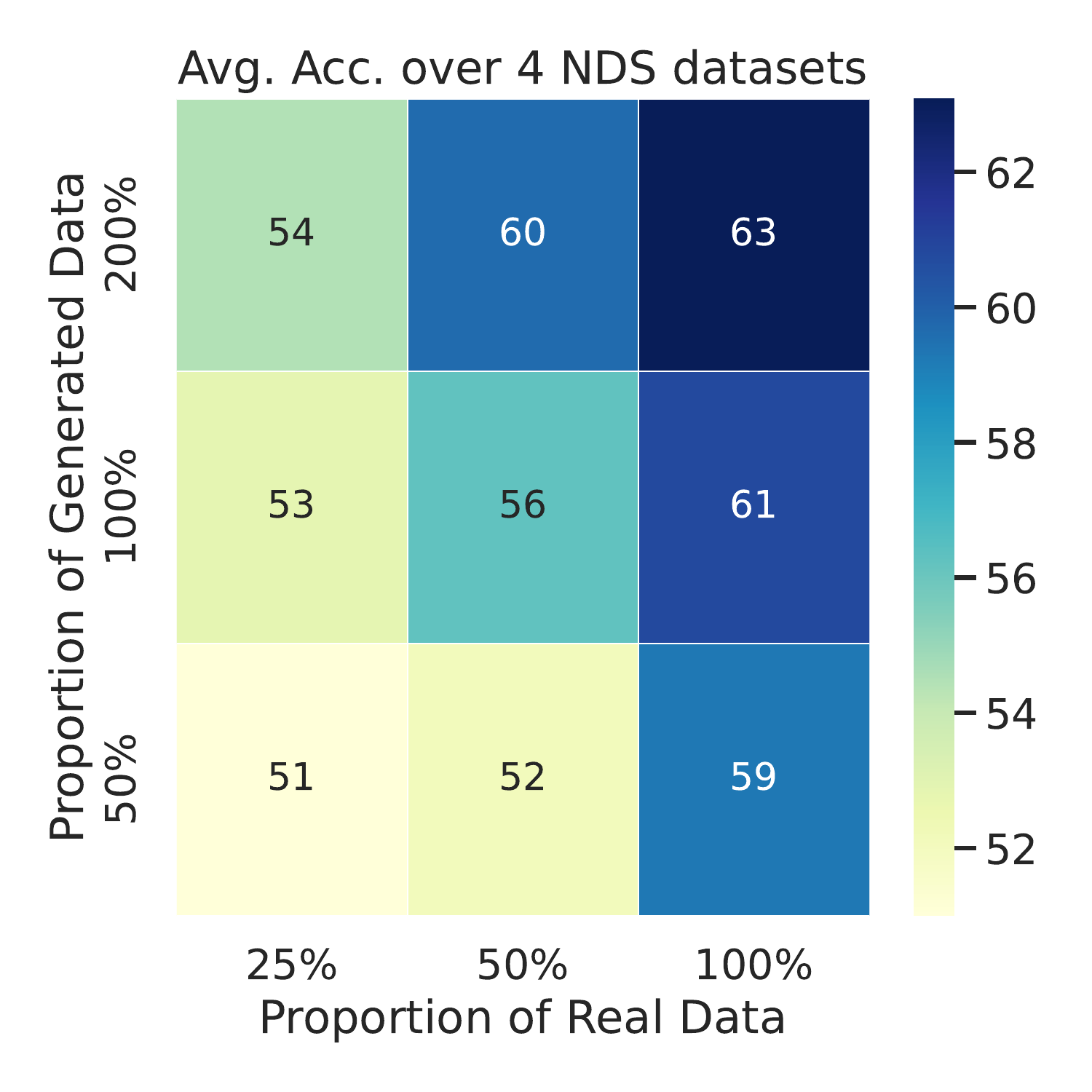}}}
    \subfloat[\centering\label{exp_fig:av_eff} Average Effective Robustness]{{\includegraphics[width=0.4\linewidth]{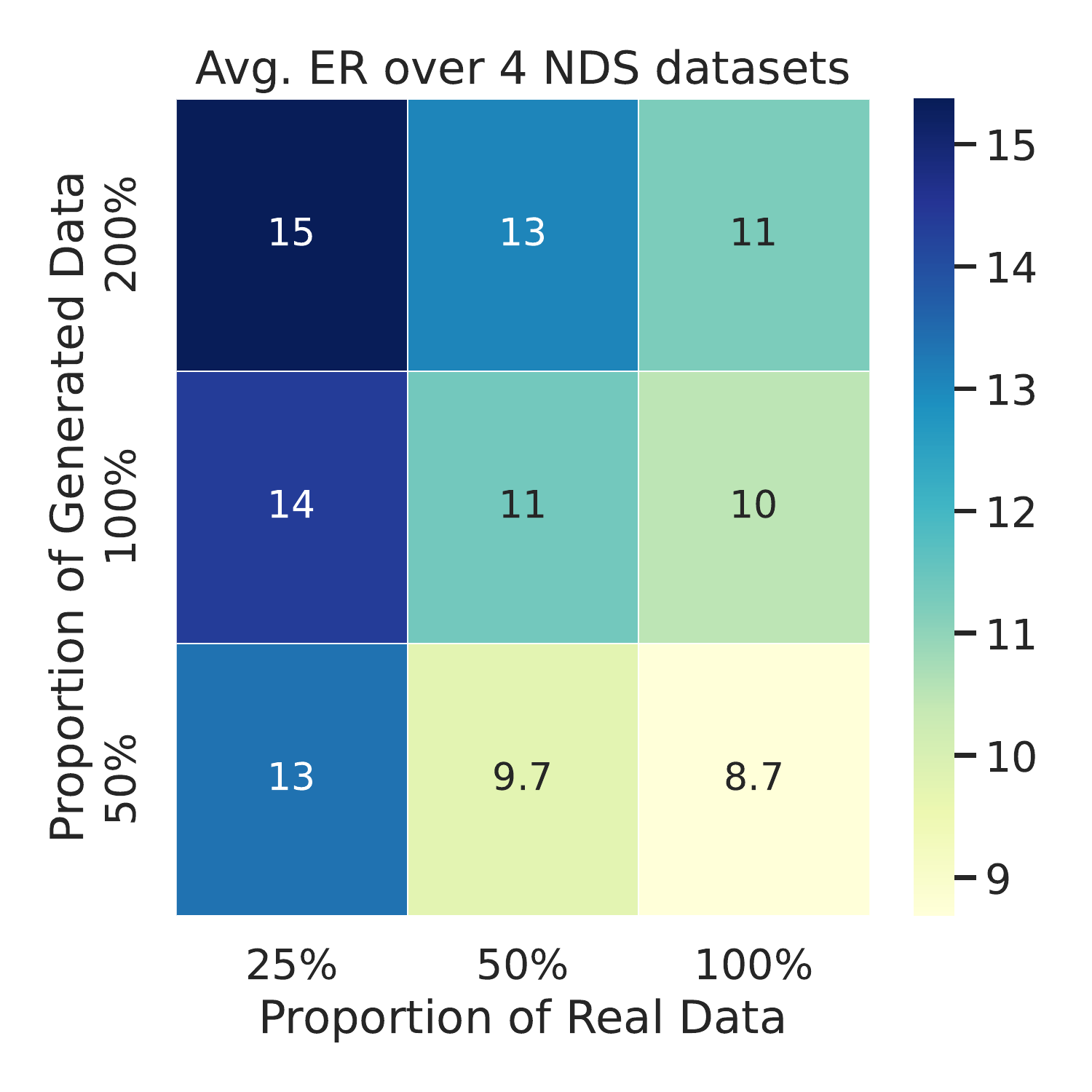}}}
    \caption{Accuracy and the Effective robustness as we vary the proportion of the real ImageNet-100 data and the generated data created using its class labels. Here $100\%$ refers to 130K training size. While calculating effective robustness, standard training is performed on $100\%$ real data.}
    \label{exp_fig:variation_in_data_size}
\end{figure}

\subsection{Effect of Real and Generated Dataset Size}
\label{exp:training_size}

% In our previous experiment in \S \ref{exp:robustness}, we observed that training a classifier on a combination of the real dataset and the generated dataset leads to improved robustness. However, to fully understand the benefit of using both datasets, we need to

Here, we investigate how different combinations of the real dataset and the generated one can help the classifiers take advantage of the complementary strengths of the two data sources. To do so, we assessed the average performance of classifiers (ResNet-18, ResNext-50, and ResNext-101) trained with six different input mixing combinations created by using ${25\%, 50\%, 100\%}$ of the real data and ${50\%, 100\%, 200\%}$ of the generated dataset using the class labels for ImageNet-100. 

As shown in Figure \ref{exp_fig:av_acc}, we observed an increase in accuracy on shifted datasets as the size of the real data increases while keeping the amount of generated data fixed. Similarly, when the proportion of the generated data increases while keeping the proportion of the real data fixed, we observed similar results. Overall, we found that increasing the amount of training data from either distribution leads to an improvement in performance on the shifted test beds. In Figure \ref{exp_fig:av_eff}, we present the average effective robustness of the classifiers across NDS datasets. Interestingly, we observe that as the proportion of generated data increases while keeping the amount of generated data fixed, the effective robustness of the classifier increases. We find that these trends remain consistent for the individual datasets in Appendix \ref{appen:variation_train_size}. In Appendix Figure \ref{appen_exp_fig:fixed_data}, we study the average trends for the accuracy and effective robustness with fixed amount of training data on ImageNet-1K.

\subsection{Effectiveness of Generation Strategies}
\label{exp:gen_strategy}

In the previous sections, we focused on generated data using 80 diverse templates with class label information from the ImageNet datasets. Here, we compare the performance of the classifiers that are trained on the real data augmented with the generated data created through mechanisms i.e., (a) diverse templates for class labels, (b) single template for class labels such as `\textit{a photo of a \textbf{class label}}', (c) real (source) images used for conditioning the generative model, and (d) real (source) images are first encoded and then denoised conditioned on the class labels. 

We report the results for ImageNet-100 in Table \ref{exp_table:generation strategies}. We find that the performance on training with synthetic dataset generated using diverse templates for class labels, or the one generated using both class labels and source images, are closely tied at $\sim$$66\%$. We observe that there is no additional benefit of using source domain information over just using the class labels information for zero-shot data generation from the modern generative models. This is different from previous works \cite{trabucco2023effective} which learns an optimized conditioning embedding from the source data to reduce the domain gap.

Further, we observe that training on the generated datasets created solely with single templates while utilizing class information results in lower robustness than training on images created via diverse templates. Interestingly, we find that the classifiers trained with images generated via a single template `a photo of a [class label]', which does not prompt the model to generate either sketches or renditions explicitly, significantly outperform the classifiers trained solely on the real data (Row 1 and Row 2). This indicates that in some cases the classifiers augmented with the generated data can be robust to specific domains without any customization during data generation. Though we lack the resources for this type of study, future work should perform large-scale human evaluations for the generated datasets along these dimensions.

\subsection{Evaluating Classifiers on Generated Datasets}
\label{exp:imnet_g_eval}

In the past sections, we established a case for using the generated data for training robust classifiers. However, the generated data can also be utilized for guiding the creation of robust image classifiers. To that end, we compare the performance of a diverse set of classifiers, (a) ResNeXt-101 trained solely on the real ImageNet-1K (ImageNet-1K), (b) ViTs pretrained on a larger set of ImageNet categories (ImageNet-21K/12K) and finetuned on ImageNet-1K, (c) Zero-shot CLIP, (d) CLIP finetuned on the real ImageNet-1K dataset, in Table \ref{exp_table:imnet_gen_eval}. We report the results of the classifiers on the original real/generated datasets, and their filtered versions that are constructed by removing all the images whose cosine similarity score with their class label's proxy caption (`a photo of a \{class label\}') is less than 0.3, as done in \cite{schuhmann2022laion}.

\begin{table}[h]
\begin{center}
\caption{Comparison of different classifiers on the original and filtered real and generated data. The accuracy gap between the performance is reported inside the gray brackets. We abbreviate Stable Diffusion as SD, Labels as L, Images as I, Pretraining as PT, \& Finetuning as FT.}
\resizebox{0.8\textwidth}{!}{%
\begin{tabular}{lcccc}
\hline
\multirow{2}{*}{\textbf{Models}} & \multicolumn{2}{c}{\textbf{Original}} & \multicolumn{2}{c}{\textbf{Filtered}} \\
\cline{2-3} \cline{4-5} 
 & Real  & Generated  & Real & Generated \\\hline
ResNeXt-101 (Real ImageNet-1K)            &         79.3             &    55.9  \textcolor{gray}{(-23.4)}                     &     90.8 & 73.2 \textcolor{gray}{(-17.6)}   \\
ViT-L/14-336 (PT-Im12K-FT-Im1K) \cite{dosovitskiy2020image} & \textbf{88.5} & 66.2 \textcolor{gray}{(-22.3)}  & 94.4 & 82.3 \textcolor{gray}{(-12.1)}  \\
MaxViT-XL-512 (PT-Im21K-FT-Im1K) \cite{tu2022maxvit} & 88.3  & 68.6 \textcolor{gray}{(-19.7)} & \textbf{94.5}  &  79.9 \textcolor{gray}{(-14.6)}  \\
Finetuned CLIP-B/32 (Real ImageNet-1K) \cite{wortsman2022robust}           &         81.3             &    64.1 \textcolor{gray}{(-17.2)}                      &    90.7 & 78.4 \textcolor{gray}{(-12.3)} \\\hline
Zero-shot CLIP-B/32 \cite{radford2021learning}           &         68.3             &    71.9  \textcolor{gray}{(+3.6)}                     &      83.1 &  85.6 \textcolor{gray}{(+2.5)}  \\
ResNeXt-101 (Real + Generated ImageNet-1K) & 80.4 & \textbf{89.0} \textcolor{gray}{(+8.6)}   &91.0 & \textbf{97.0} \textcolor{gray}{(+6.0)} \\\hline
\end{tabular}%
}
\label{exp_table:imnet_gen_eval}
\end{center}
\end{table}

Despite performing the best on ImageNet-1K validation datasets, ViTs underperform on the generated data. We further find that the CLIP finetuned on ImageNet-1K experiences a performance degradation of upto $17\%$, $12\%$ absolute accuracy on the original and filtered datasets respectively. However, we find that zero-shot CLIP does not undergo a distribution shift on the generated data. Since the zero-shot CLIP encoders are used as module in our data generator Stable Diffusion, the good performance of CLIP on the generated dataset underscores a ``cyclic consistent" nature where the conditional generations of an encoder-decoder generative model (Stable Diffusion) agree with the standalone encoders in CLIP. To better quantify the performance gap on the generated data, we evaluate the performance of a classifier trained on the combination of the real and generated data. We observe that the classifier achieves upto $89\%$, $97\%$ on the real and generated data, respectively, which highlights the potential for further improvements of the existing models on the novel realizations of the ImageNet objects.

\section{Generated Data from Finetuned Stable Diffusion}
\label{sd-finetune-im1k}

In our work, we showed that the classifiers trained on the real data augmented with the generated data, acquired in a zero-shot manner from the base generative model, are robust to natural distribution shifts. Here, we aim to study the impact of varying the data generation paradigm and evaluate the quality of the image classifiers trained on the generated data that is closer in distribution to the real data as compared to the generated data collected in a zero-shot way.

To this end, we finetune the base Stable Diffusion v1.5 for 1 epoch on the complete 1.3M (real) ImageNet-1K data and their corresponding class labels, at the default resolution of 512 x 512.\footnote{Our finetuning recipe along with the checkpoint is available at \url{https://huggingface.co/hbXNov/ucla-mint-finetune-sd-im1k}} Post-finetuning, we repeatedly query the generative model conditioned on the class labels to synthesize a newly generated data of the same size as ImageNet-1K training and validation datasets. Finally, we train ResNext-50 classifier (a) solely on the newly generated data, and (b) an equal mix of real data and newly generated data, from the finetuned Stable Diffusion. In Table \ref{appen_table:comparison_base_finetuned_sd}, we compare the performance of the same classifier trained with the (a) real data, (b) generated data from the base generative model conditioned on the class labels, and (c) an equal mix of the real and base generated data, on the real ImageNet-1K test set and its natural distribution shift datasets.

\begin{table}[h]
\begin{center}
\caption{Comparison of the performance of a ResNext-50 classifier on the ImageNet-1K validation dataset, and its natural distribution shift datasets. The training data contains 1.3M examples for the Real, Base-Generated, and Finetune-Generated data. Here, Real + Base-Generated or Finetune-Generated indicates that the generated data is used to augment the real data.}
\resizebox{\textwidth}{!}{%
\begin{tabular}{lccccc|c}
\hline
\textbf{Data}          & \textbf{ImageNet} & \textbf{ImageNet-Sketch} & \textbf{ImageNet-R} & \textbf{ImageNet-V2 }& \textbf{ObjectNet} & \textbf{Average}\\\hline
Real                & 78.4     & 25.0          & 42.2     & 68.5      & 40.6 &  51.0 \\
Base-Generated            & 32.4    & 21.6       & 37.4    & 26.2     & 19.4  & 27.4\\
Finetune-Generated        & 38.1     & 9.4          & 18.4 & 28.0         & 16.7 & 22.1 \\\hline
Real + Base-Generated     & 78.4    & 40.1          & 56.2 & 66.5    & 39.4  & \textbf{56.1}\\
Real + Finetune-Generated & 78.0    & 28.2          & 41.5     & 66.0      & 37.5 & 50.2  \\\hline
\end{tabular}%
}
\label{appen_table:comparison_base_finetuned_sd}
\end{center}
\end{table}

We find that the image classifiers trained with solely the finetuned-generated data (Row 3) outperform the one trained with the base-generated data (Row 2) on the ImageNet-1K validation dataset. This is due to the reduction in the distribution gap between the real data and the generated data from the finetuned Stable Diffusion model. We note that the accuracy achieved by the classifiers trained on the finetuned Stable Diffusion i.e., $38.1\%$ lags behind the accuracy achieved in \cite{azizi2023synthetic} by training on the generated data from the finetuned ImaGen model i.e., $67\%$. We attribute this difference in the accuracies to the differences in the quality of the base generative models themselves. 

Despite the reduction in the domain gap between the real data and generated data via finetuning, we find that the ImageNet-1K validation accuracy for the classifier trained on the real data augmented with the finetuned generated data $78\%$ (Row 5) is close to the one trained on the real data augmented with the generated data from the base model $78.4\%$ (Row 4). Although our observation may surprising, we find that similar observations were made in Table 4 in  \cite{azizi2023synthetic} and Figure 5 in \cite{ravuri2019classification} at high resolutions. The exact reason behind this empirical finding is still unclear, and a potential future work. 

Lastly, we observe that the accuracy gains over standard training on the natural distribution datasets are higher for the classifier trained on the real data augmented with the base-generated data as compared to the one trained on the real data augmented with the finetuned generated data. For example, the classifier trained on the real and base-generated data achieves an accuracy of $40.1\%$ and $56.2\%$ whereas the classifier trained on the real and finetuned-generated data achieves an accuracy of $56.2\%$ and $41.5\%$ on ImageNet-Sketch and ImageNet-R, respectively. Our finding further highlights the usefulness of training the classifiers on the diverse data, from the base generative model, over the generated data that is closer to the real data distribution, on natural distribution shift datasets. 
\section{Related Work}
\label{relatedwork}

\textbf{Training Robust Classifiers:} 
% The goal of machine learning is to train neural classifiers that can generalize beyond the training distribution. Data augmentation and pre-training on diverse, in-the-wild data have been shown to be effective in increasing the robustness of classifiers. 
Many works propose hand-engineered augmentations to increase the training data and improve generalizability of the classifiers, e.g., \cite{hendrycks2019augmix,hendrycks2022pixmix,zhang2017mixup}. \cite{cubuk2018autoaugment,cubuk2020randaugment} learn augmentation policies directly from the data and have been shown to improve classification accuracy. DeepAugment \cite{hendrycks2021many} was one of the first augmentation strategies to perform well on natural distribution shifts. Additionally, studies on CLIP-verse \cite{radford2021learning,jia2021scaling,li2021supervision,goel2022cyclip,mu2022slip} have shown natural robustness. In our work, we take the best of both paradigms by leveraging the strengths of modern generative models to augment real datasets. We find that classifiers trained with generated datasets are effectively robust and outperform current data augmentation strategies in eliciting robustness.

\textbf{Robustness via Generated Data:} 
% In recent years, the ability to generate high-fidelity images at accelerated speeds has revolutionized generative modeling \cite{brock2018large,child2020very}. 
\cite{gowal2021improving,sehwag2021robust} studied the effectiveness of synthetic data from these models for creating adversarially robust classifiers, but did not examine the robustness in the regime of natural distribution shifts (NDS) and modern in-the-wild generative models \cite{rombach2022high,ramesh2021zero,xu2022versatile,saharia2022photorealistic,balaji2022ediffi,chang2023muse}. \cite{he2022synthetic} generates synthetic data using the GLIDE \cite{nichol2021glide} and finds that it improves the accuracy of the CLIP model \cite{radford2021learning}, indicating the usefulness of synthetic data for pre-training image models. Our work focuses on the use-case of the generated data, created in a zero-shot manner, for training robust image classifiers against natural distribution shifts, and benchmarking the existing image classifiers. 

% However, we perform a detailed analysis of the effectiveness of generated data for robust classification, specifically focusing on natural distribution shift datasets. \cite{yuan2022not} adapt to the target domain by training on a generated dataset using the variations of the images in the source domain. In contrast, our work does not make assumptions about the target domain and does not require access to source images to train robust classifiers. Our study further includes experiments on the effectiveness of various generation strategies to elicit robustness to NDS for larger datasets such as ImageNet.

\textbf{Model Evaluation:} 
% As new models are developed with advancements in model and training size, inductive biases, and training procedures, it is important to benchmark their progress against standard benchmarks. 
% Several studies \cite{wang2022generalizing,wang2018deep,du2020learning,blanchard2021domain} have evaluated the ability of image classifiers to generalize to unseen domains after being trained on a source domain. Other works focus on evaluating a model trained with a source distribution on a \textit{related} but different distribution. In this paradigm, \cite{hendrycks2019benchmarking} showed that ImageNet models are sensitive to synthetic corruptions such as noise, blur, and compression in the input images. 
Studies by \cite{recht2018cifar,recht2019imagenet,hendrycks2021many,wang2019learning,barbu2019objectnet} assess the model's ability to generalize to natural variations in images containing objects from the source dataset, showing severe performance dips and questioning their usefulness for real-world applications. In our work, we create a generated validation set from a modern generative model, containing new realizations of the objects in the ImageNet-1K dataset that may be difficult to acquire in the real-world. We find that the state-of-the-art ImageNet classifiers experience a performance degradation on the generated validation data, highlighting at a gap that the robustness research should aim to bridge.

\textbf{Augmenting with Generated Data:} \cite{antoniou2017data} used generated data to enhance the diversity of training data, leading to improved classification results, via an image-conditional GAN \cite{goodfellow2020generative}. Since then, numerous studies have applied generated data in various domains. \cite{west2021symbolic} generated a massive commonsense knowledge corpus using GPT-3 \cite{brown2020language} to train commonsense models. Brooks et al. \cite{brooks2022instructpix2pix} fine-tuned a stable diffusion model with a set of creative image-text pairs generated from a combination of GPT-3 and Stable diffusion for image editing. Our work demonstrates a practical application of using generated data for improved robustness in model training.
\section{Conclusion}
\label{conclusion}

We developed a framework to improve performance of image classifiers by augmenting real datasets with a diverse dataset generated by a modern `in-the-wild' generative models. Our results show that classifiers trained with this method exhibit high performance on test and natural distribution shift datasets. This is due to the increased robustness obtained from training on generated data compared to standard training methods. We also analyzed the role of different generation strategies to better explain these trends. Additionally, we used the synthetic data as an evaluation dataset and highlighted the brittleness of state-of-the-art models to natural variations in generated images. Finally, we showed that the generated data from the base generative model has more practical usefulness for training robust classifiers as compared to the generated data from a finetuned generative model on the real ImageNet data. A current limitation is evaluating the trustworthiness of generated data based solely on robustness. Future research should incorporate a multi-dimensional analysis, including factors such as privacy and the presence of harmful stereotypes. The total computational cost of our framework includes the cost of creating a generated dataset, and of training the classifiers on the real data augmented with the generated data. Though we lack the resources for this type of study, future work should also investigate scaling laws for generated datasets. Finally, it would be compelling to perform large-scale human annotations for a better understanding of the failure modes of the generated datasets.

\bibliography{main}
\bibliographystyle{plain}

\newpage

\appendix

\section{Limitations}
\label{limitations}

While the ability to train robust classifiers using generated data from modern text-to-image generative models represents a significant advancement in generative AI for trustworthy machine learning (ML), there are other equally important aspects, such as fairness and privacy, that have not been explored in this work. In this study, our focus is on highlighting the benefits of generated data for objects in the ImageNet-1K dataset. However, this raises intriguing questions about the generalizability of these results to larger datasets like ImageNet-21K \cite{ridnik2021imagenet}.

Our approach primarily concentrates on generating data from the base generative model in a zero-shot manner for objects that are well-represented within its distribution. Nevertheless, it is crucial to fine-tune the base model for domains that are not adequately captured in its training distribution, such as medical images \cite{chambon2022adapting}. Despite these limitations, the core contributions of this paper remain highly valuable and provide crucial insights for promoting positive impact in trustworthy ML.

\section{Broader Impact}
\label{ethics}

In our work, we utilize modern `in the wild' generative models to create generated data, that is further employed for training Image classifiers. Since these generative models are trained on large, diverse, and uncurated web-scraped datasets, there are several privacy concerns surrounding the suitable use of public data \cite{scheuerman2021datasets}, and their harmful biases and stereotypes \cite{birhane2021multimodal,bender2021dangers}. Once trained, these generative models can amplify these biases during generation \cite{saharia2022photorealistic,cho2022dall,bansal2022well}. With the generative model's ability to create and combine different concepts in realistic ways, there are harms associated with changing the predictions based on the natural language descriptions of the concepts as it is much easier to generate objectionable content with these. It necessitates further research into closely curating the generated data as well as building fairer multimodal representations of the real world.

As generated data pervades the Internet, it is inevitable that they will be explicitly used or automatically scraped as training data for building new data-driven models, such as our work. These scenarios present a difficult challenge for researchers to better understand and track the source of harmful biases introduced in the dataset. Additionally, there are equally relevant privacy concerns as we train on the model generations, which in recent times, have been shown to replicate styles of real artists \cite{cooper2022}. Hence, making the generated dataset publicly available is a step in the direction towards future benchmarking and critique of the design and use of generated datasets for trustworthy ML.

\section{Background - Data Generation using Stable Diffusion}
\label{appen_data_gen_sd}

In this work, we employ Stable Diffusion (SD) \cite{rombach2022high}, an `in the wild' generative model is one that can generate images from the natural language description of a wide range of concepts, combine unrelated concepts in a realistic manner, and apply novel transformations to existing images. Such abilities are exhibited by Stable Diffusion through training on a large, diverse dataset LAION \cite{schuhmann2022laion} on matched image-text pairs $(\mathcal{X}, \mathcal{C})$ scraped from the web where $\mathbf{x} \subset \mathcal{X}$ denotes a raw image and $c \subset \mathcal{C}$ denotes its corresponding caption in natural language. 

During training, the image $\mathbf{x}$ is passed through a pre-trained encoder $z_0 = \mathcal{E}(\mathbf{x})$ where $z_0$ is the latent representation of $x$. The objective of the denoising model $R(z_t, t, y_g)$ is to predict $z_0$ from every intermediate representation $z_t$ where $z_t$ is sampled from  $t:=1,\ldots,T$ while the conditioning variable $y_g$ guides the training process. For image generation, we sample from $z \sim N(0, I)$, and use the trained model $R(.)$ with a predefined sampling scheme (DDPM \cite{ho2020denoising}, DDIM \cite{song2020denoising}) to reconstruct $z_0$ iteratively. Finally, the latent representation $z_0$ is decoded using the pretrained decoder $\mathbf{x}_g = \mathcal{D}(z_0)$ to generate the synthetic image $\mathbf{x}_g$. 

Given a single data point $(\mathbf{x}, y)$ from the source dataset, we have various ways to generate a new data point $\mathbf{x}_g$ with a trained Stable Diffusion, as summarized in Appendix Figure \ref{appen_fig:gen_strategies}.

\textbf{Generation via Class Labels}: In practice, Stable Diffusion uses CLIP's \cite{radford2021learning} text encoder $y_g = h_{text}(c)$ for conditioning during the training process. Here, we synthesize images by denoising $z_T \sim N(0, I)$ conditioned on the natural language templates $\mathcal{M}$ for the class labels $y$. An example template $M \subset \mathcal{M}$ includes `A photo of a \textit{dog}' where \textit{dog} is the class label $y$. This generation strategy involves using a pretrained CLIP text encoder $y_g = h_{text}(M(y))$. Since generating data conditioned on the natural text descriptions is the default setting for data generation using Stable Diffusion, our primary focus is on the natural robustness elicited by this data generation strategy.

In addition to the traditional zero-shot data generation approach, we study the following other ways to generate images without any training or finetuning of the generative model on the images from the source dataset. We specifically study the effect of these data generation procedures in \S \ref{exp:gen_strategy}. 

\textbf{Generation via Real (Source) Images:} Here, we use CLIP's vision encoder $y_g = h_{image}(\mathbf{x})$ for conditioning. In this case, we generate variations of the images from the source dataset by denoising the latent variable $z_T$ conditioned on their representations. 

\textbf{Generation via Real (Source) Images and Class Labels:} We can create realistic variations of the source image $\mathbf{x}$ by first encoding it using the pretrained encoder $\mathcal{E}(\mathbf{x})$ followed by forward diffusion for $T$ steps to approximate a normal distribution $\hat{z}_{T}(\mathbf{x})$. Consequently, we generate a new image by denoising $\hat{z}_{T}(\mathbf{x})$ conditioned on the natural description of the class label $y_g = h_{text}(M(y))$.

\begin{figure}[h]
\centering
\subfloat[\centering \label{exp_fig:av_acc} Data Generation using either the proxy captions for class labels or the source image from the dataset.]{{\includegraphics[width=0.475\linewidth]{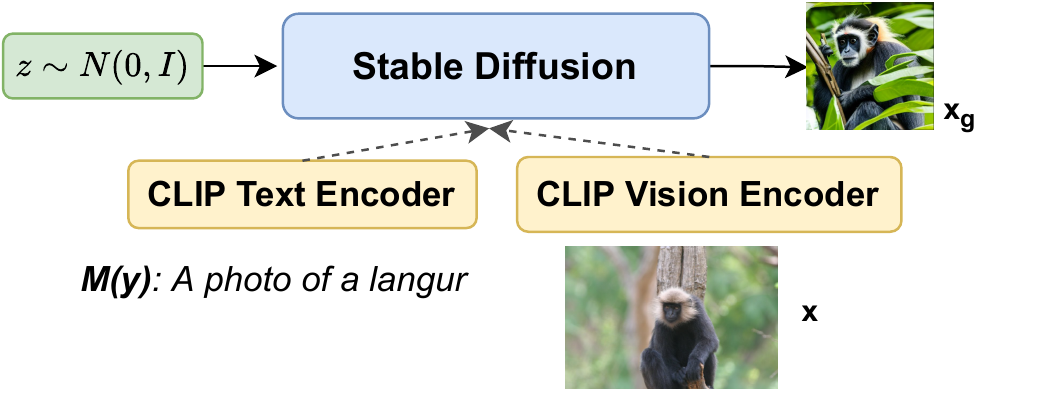}}}%
\subfloat[\centering\label{exp_fig:av_eff} Data Generation using both the proxy captions and the source image from the dataset.]{{\includegraphics[width=0.475\linewidth]{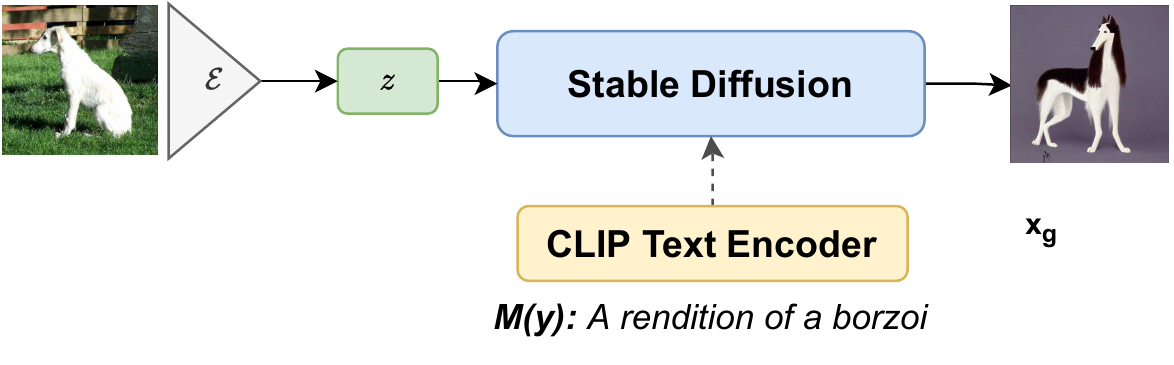}}}%
\caption{Overview of our generation strategies. We use Stable Diffusion (SD) to create the generated dataset. (a) We can create images by conditioning on either the proxy caption for the class label (Generation via Class Labels), or conditioning on the images from the source dataset (Generation via Real Images). (b) We can also generate data by first encoding the source images to get the latent representation, which is then denoised conditioned on the text prompt for the class label (Generation via Real Images and Class Labels).}
\label{appen_fig:gen_strategies}
\end{figure}

\section{Setup}
\label{appen:setup}

It took us $\sim$10 days to generate the complete dataset on 5 Nvidia RTX A5000 GPUs with a batch size of 12 per GPU. Additionally, we generate a separate training dataset of 130K images and a validation dataset of 5K images for every generation strategy described in \S \ref{background:sd}. We present some sample generations in Appendix Figure \ref{appen_fig:generation_visualization}.

\section{Generated Data Analysis}
\label{appen:human_eval}

\begin{table}[h]
\caption{Comparison of consistency (0-1) and quality (1-5) between the real images and the synthetic images created using various generation images. The numbers are averaged over the individual scores of the 20 human annotators.}
\begin{center}
\resizebox{\textwidth}{!}{%
\begin{tabular}{lcccc}
\hline
            & Real & Generated (\small{Class Labels}) & Generated (\small{Real Images}) & Generated (\small{Real and Class Labels}) \\\hline
Consistency (Humans) & \textbf{0.96}       & 0.86   & 0.54 & 0.85               \\
Quality (Humans)    & \textbf{4.52}      & 4.2   & 2.96 & 3.8           \\
Diversity (CLIP) & \textbf{0.30} & 0.26& 0.32 & 0.23\\\hline
\end{tabular}%
}
\end{center}
\label{exp_table:data_analysis}
\end{table}

Since the generative model is prompted in a \textit{zero-shot} manner, it is important to compare the consistency, quality, and diversity of the generated data with the real data. To do so, we perform a human evaluation study to assess whether there is a lack of useful information in the generated datasets that might be relevant to classify an object (Consistency), and whether the generated images are of poor quality i.e., they lack sharpness or contain perceptible noise (Quality). We collect 1600 annotations from 20 human surveyors for 40 images that are sampled from different real/generated datasets from 10 ImageNet classes. Further details on the data collection process are presented in Appendix \S \ref{appen:human_eval}. In addition, we compare the diversity in the real and generated dataset by subtracting the average of 1 - mean cosine pair-wise similarities between the CLIP representations of the images within each class of ImageNet-100, as done in \cite{udandarao2022sus}.

We find that images belonging to the real ImageNet dataset are more consistent, of higher quality and more diverse than generated data created by conditioning a modern generative model on the natural descriptions of the class labels. This is expected since the real ImageNet went through extensive data curation and cleaning process during its creation. Since the scores for the generated data via class labels are not that far off, it provides further evidence for its effectiveness and potential training robust classifiers. In addition, we observe that the consistency and quality scores of images generated via class labels and the ones generated via source images and class labels are close. In terms of the diversity, we observe that data generation
using only source image information led to the
most diverse creations within each class. However, we also find that synthetic data generated using just the source images had low consistency and quality scores, suggesting at the poor object representations and image quality, which do not aid in robustness to natural distribution shifts.

\subsection{Human Evaluation}
\begin{figure}[h]
    \centering
    \includegraphics[width=0.5\linewidth]{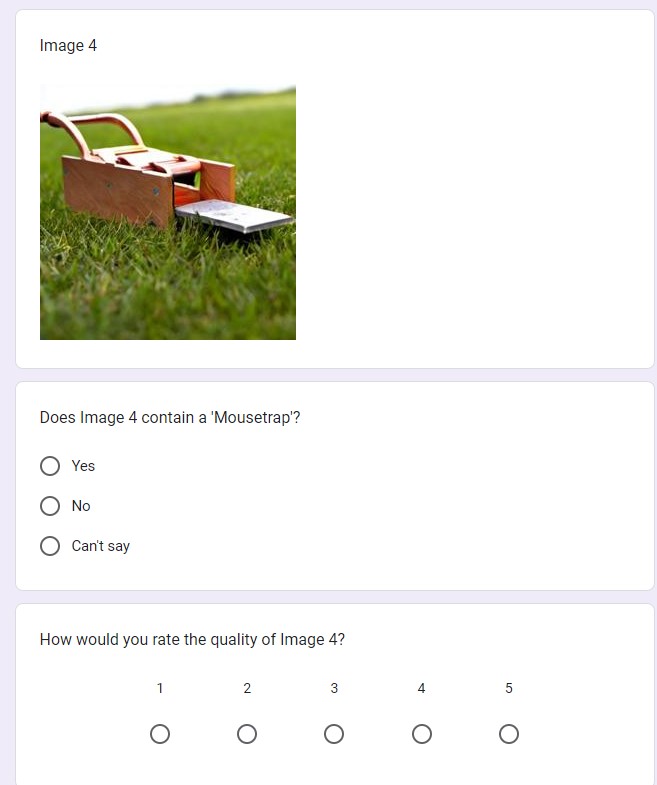}
    \caption{Survey screenshot}
    \label{appen_fig:survey_ss}
\end{figure}

We randomly selected images from 10 classes of the ImageNet1K dataset and used them to synthesize generated images using three different strategies: generated data via class labels, via real (source) images, and a combination of both, as described in \S \ref{exp:gen_strategy}. This resulted in a total of 40 images for our study, including the real images. We then recruited a pool of 20 human annotators to independently complete a survey in which they were shown each image without any information about its source. \footnote{Human annotators are graduate students from the department of CS at UCLA.} They were asked two questions for each image: 1) whether the image contained the intended class label, and 2) to rate the image's quality on a scale of 1-5. The screenshot of the survey for one image is provided for reference in Figure~\ref{appen_fig:survey_ss}.

\section{Setup for Training Image Classifiers}
\label{appen:setup_train_classifier}

As suggested in previous studies \cite{kusupati2022matryoshka}, we train all the models using the efficient dataloaders of FFCV \cite{leclerc2022ffcv}. We train the models for 40 epochs with the batch size of 512 on ImageNet-1K, and for 88 epochs with the batch size of 512 on ImageNet-100. All the models are trained with a learning rate of 0.5 with a cyclic learning rate schedule \cite{smith2017cyclical}. All the models are trained with SGD optimizer with a weight decay of 5e-5.

\section{More Details on Natural Distribution Shift Datasets}
\label{appen:nds}

ImageNet-Sketch contains the sketches of ImageNet-1K objects. ImageNet-R contains the renditions (paintings, sculptures) for 200 ImageNet-1K classes, 19 of which overlap with ImageNet-100. ImageNet-V2 is a reproduction of ImageNet-1K validation dataset, and we consider its matched frequency variant that closely follows the ImageNet-1K data distribution. Finally, ObjectNet contains a objects in novel backgrounds and rotations with 113 overlapping classes with ImageNet-1K, and 13 classes overlapping with ImageNet-100.

\section{Templates used for Data Generation}
\label{appendix:templates}

We present the list of 80 diverse templates that were used to generate the new images in Table \ref{appen_table:templates}.

\begin{table}[h]
\begin{center}
\begin{tabular}{|p{1cm}|p{16cm}|}
\hline
 & 'a bad photo of a \{class label\}.', 'a photo of many \{class label\}.', 'a sculpture of a \{class label\}.', 'a photo of the hard to see \{class label\}.', 'a low resolution photo of the \{class label\}.', 'a rendering of a \{class label\}.', 'graffiti of a \{class label\}.', 'a bad photo of the \{class label\}.', 'a cropped photo of the \{class label\}.', 'a tattoo of a \{class label\}.', 'the embroidered \{class label\}.', 'a photo of a hard to see \{class label\}.', 'a bright photo of a \{class label\}.', 'a photo of a clean \{class label\}.', 'a photo of a dirty \{class label\}.', 'a dark photo of the \{class label\}.', 'a drawing of a \{class label\}.', 'a photo of my \{class label\}.', 'the plastic \{class label\}.', 'a photo of the cool \{class label\}.', 'a close-up photo of a \{class label\}.', 'a black and white photo of the \{class label\}.', 'a painting of the \{class label\}.', 'a painting of a \{class label\}.', 'a pixelated photo of the \{class label\}.', 'a sculpture of the \{class label\}.', 'a bright photo of the \{class label\}.', 'a cropped photo of a \{class label\}.', 'a plastic \{class label\}.', 'a photo of the dirty \{class label\}.', 'a jpeg corrupted photo of a \{class label\}.', 'a blurry photo of the \{class label\}.', 'a photo of the \{class label\}.', 'a good photo of the \{class label\}.', 'a rendering of the \{class label\}.', 'a \{class label\} in a video game.', 'a photo of one \{class label\}.', 'a doodle of a \{class label\}.', 'a close-up photo of the \{class label\}.', 'a photo of a \{class label\}.', 'the origami \{class label\}.', 'the \{class label\} in a video game.', 'a sketch of a \{class label\}.', 'a doodle of the \{class label\}.', 'a origami \{class label\}.', 'a low resolution photo of a \{class label\}.', 'the toy \{class label\}.', 'a rendition of the \{class label\}.', 'a photo of the clean \{class label\}.', 'a photo of a large \{class label\}.', 'a rendition of a \{class label\}.', 'a photo of a nice \{class label\}.', 'a photo of a weird \{class label\}.', 'a blurry photo of a \{class label\}.', 'a cartoon \{class label\}.', 'art of a \{class label\}.', 'a sketch of the \{class label\}.', 'a embroidered \{class label\}.', 'a pixelated photo of a \{class label\}.', 'itap of the \{class label\}.', 'a jpeg corrupted photo of the \{class label\}.', 'a good photo of a \{class label\}.', 'a plushie \{class label\}.', 'a photo of the nice \{class label\}.', 'a photo of the small \{class label\}.', 'a photo of the weird \{class label\}.', 'the cartoon \{class label\}.', 'art of the \{class label\}.', 'a drawing of the \{class label\}.', 'a photo of the large \{class label\}.', 'a black and white photo of a \{class label\}.', 'the plushie \{class label\}.', 'a dark photo of a \{class label\}.', 'itap of a \{class label\}.', 'graffiti of the \{class label\}.', 'a toy \{class label\}.', 'itap of my \{class label\}.', 'a photo of a cool \{class label\}.', 'a photo of a small \{class label\}.', 'a tattoo of the \{class label\}.'\\\hline
\end{tabular}
\caption{List of diverse templates used for generating data.}
\label{appen_table:templates}
\end{center}
\end{table}

\section{Visualization of Image Generations}

\begin{figure}[h]
    \centering
    \includegraphics[width=0.95\linewidth]{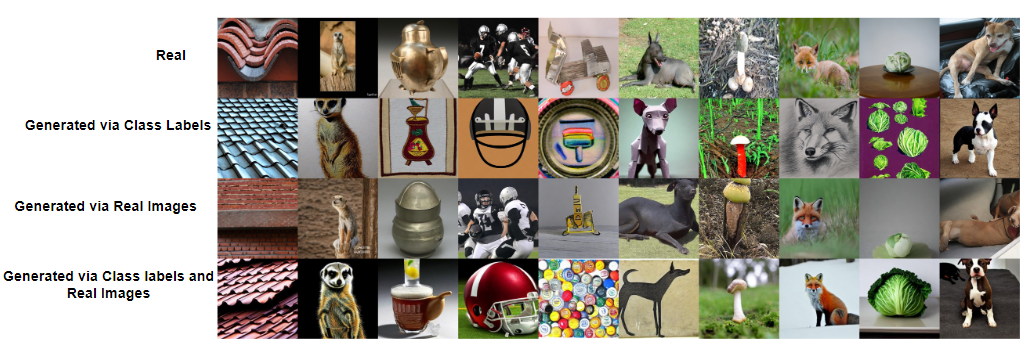}
    \caption{Visualization of samples from the real dataset and various generation strategies using Stable Diffusion (SD).}
    \label{appen_fig:generation_visualization}
\end{figure}

We present a sample visualizations of the images generated via different generated strategies in Figure \ref{appen_fig:generation_visualization}.

\section{ImageNet-C}
\label{appen:Im-C}

The evaluation datasets such as ImageNet-C intend to perturb the real images and distort their quality, such that the representations of the perturbed images are pushed outside the decision boundary of their true class ids. This differs from natural distribution shift datasets such as ImageNet-V2, ObjectNet, ImageNet-R, and ImageNet-Sketch, since these datasets are acquired under different environments in the real-world rather than formed by perturbing the original datasets themselves. To understand the usefulness of the generated data for ImageNet-C, we provide the results for the absolute accuracy and effective robustness of the models on ImageNet-C (Severity-5). We report the average accuracy over all the sub-datasets in the ImageNet-C, in Table \ref{exp:im-c}.

\begin{table}[h]
\caption{Comparison of training ImageNet-1K classifiers on the real data, generated data, and the equal mix of real and generated data, on ImageNet-C (Severity = 5) validation datasets.}
\label{exp:im-c}
\begin{center}
\begin{tabular}{lcc}
\hline
Method                     & Accuracy (\%) & Effective Robustness (\%) \\\hline

Real Data                  & 20.5          & -                         \\
Generated Data             & 3.3           & \textbf{25.5}\\
Real Data + Generated Data & \textbf{21.75}         & 1.3   \\\hline
\end{tabular}
\end{center}                   
\end{table}

We find that the classifiers trained with solely the generated data as well as the mix of real and generated achieve high effectiveness robustness over standard training on the real data (Column 2). The absolute accuracy increases by 1.25\% on the validation set of the ImageNet-1K using our augmentation.

\section{Effect of changing the training size}
\label{appen:variation_train_size}

\begin{figure}[h]
    \centering
    \subfloat[\centering\label{appen_fig:av_acc_s} Accuracy]{{\includegraphics[width=0.35\linewidth]{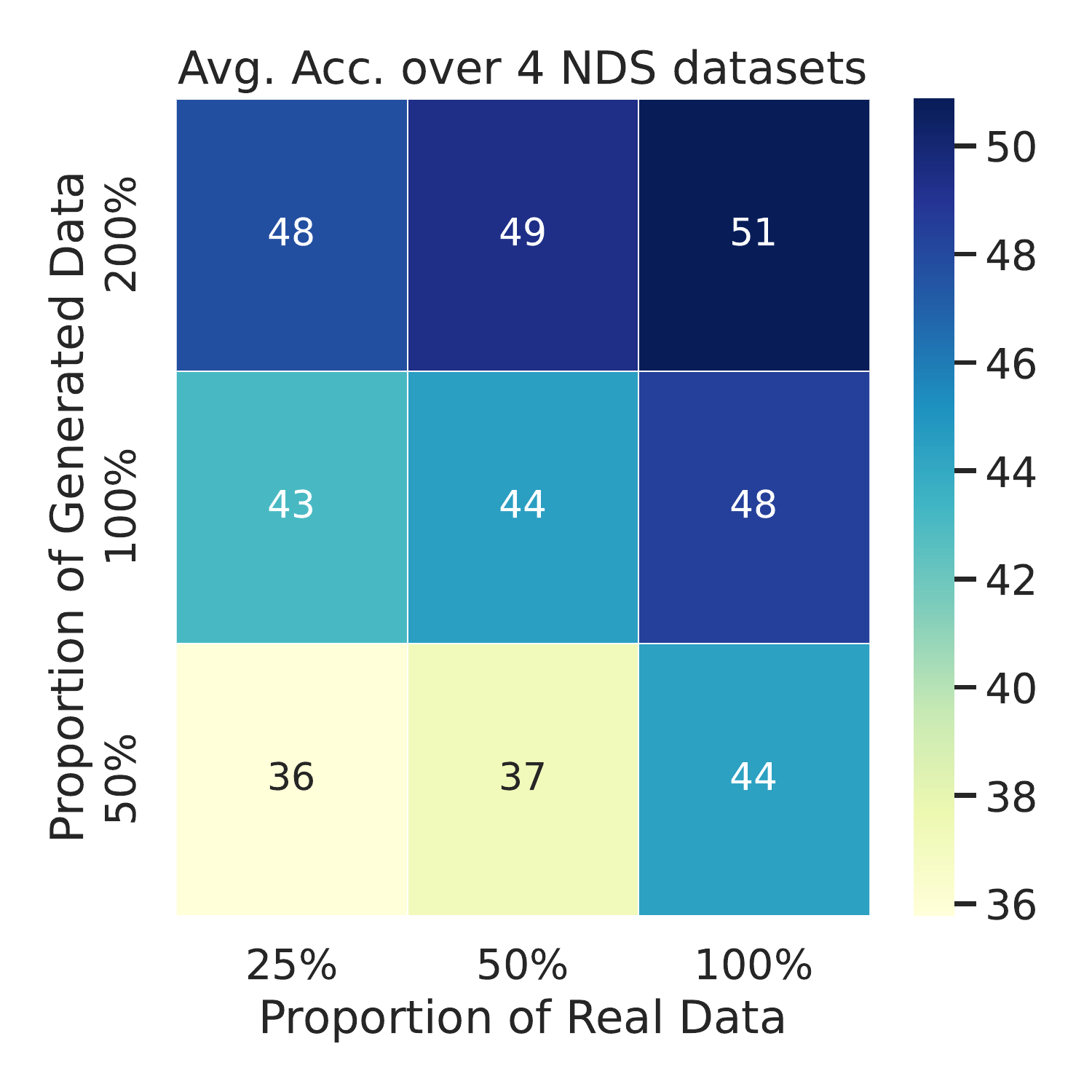}}}
    \subfloat[\centering\label{appen_fig:av_eff_s}  Effective Robustness]{{\includegraphics[width=0.35\linewidth]{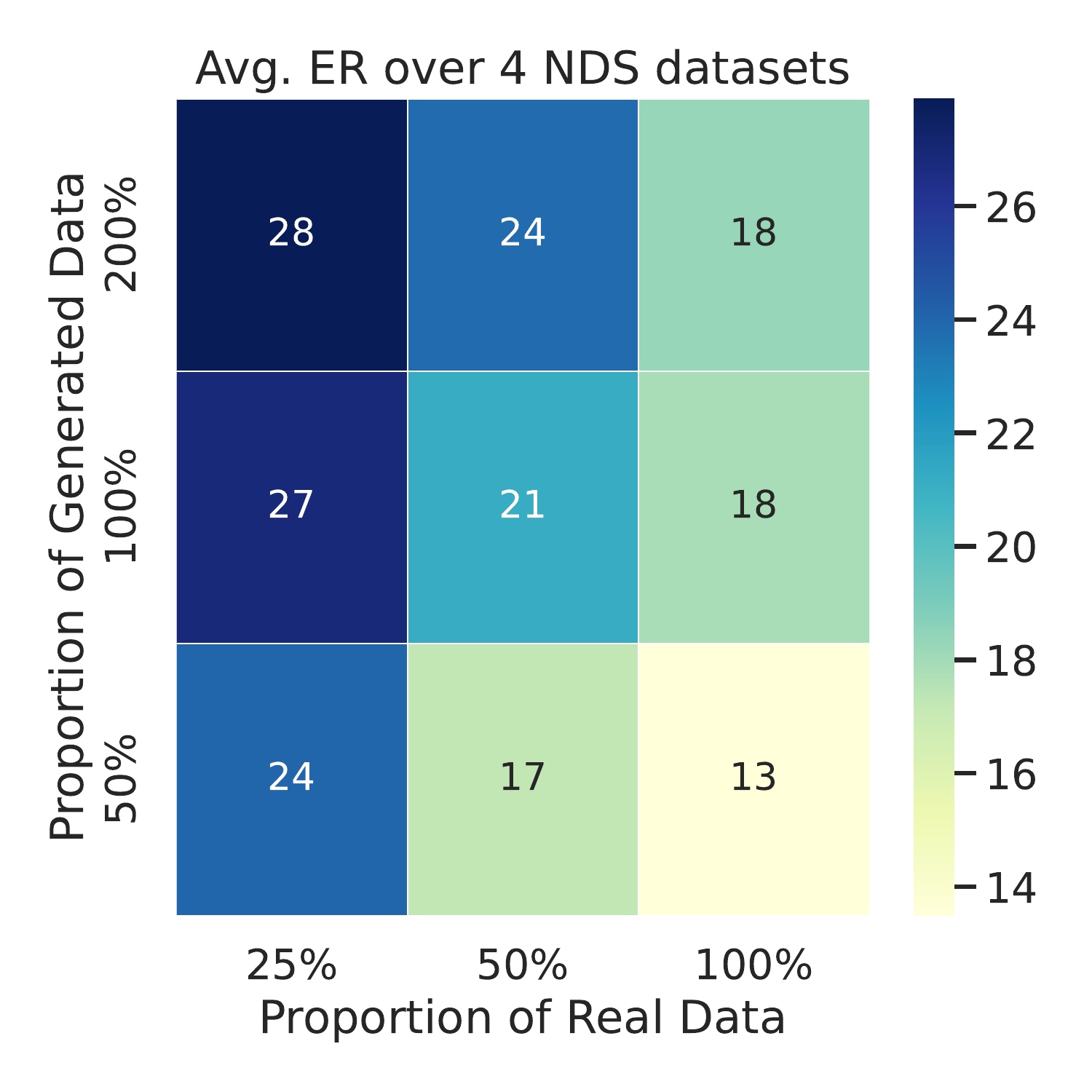}}}
    \caption{Variation in the accuracy and the effective robustness on ImageNet-Sketch as we vary the proportion of the real ImageNet-100 data and the generated data created using its class labels in the training set. Here $100\%$ refers to 130K training size. While calculating effective robustness, standard training is performed on $100\%$ real data.}
    \label{appen_fig:variation_sketch}
\end{figure}

\begin{figure}[h]
    \centering
    \subfloat[\centering\label{appen_fig:av_acc_r} Accuracy]{{\includegraphics[width=0.35\linewidth]{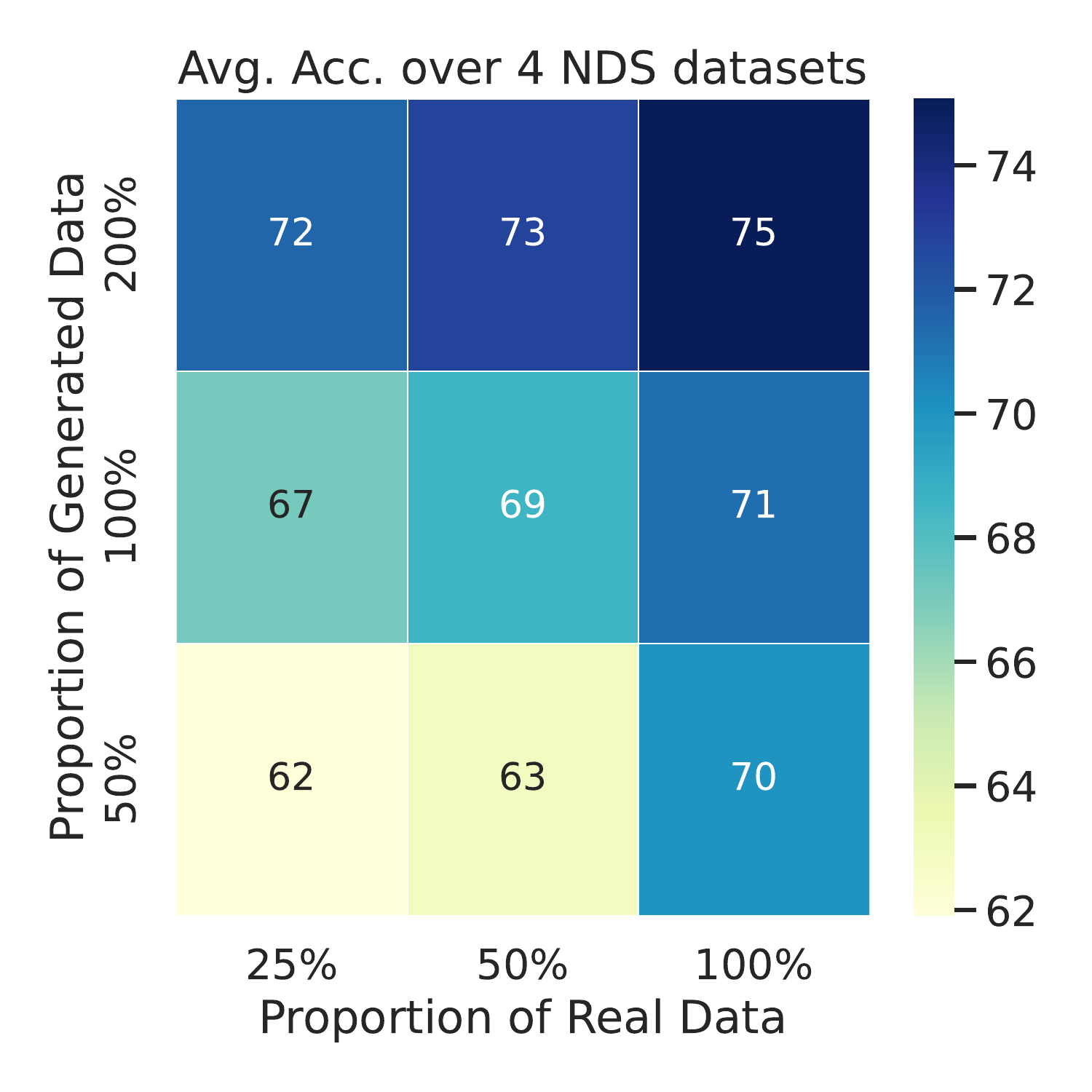}}}
    \subfloat[\centering\label{appen_fig:av_eff_r}  Effective Robustness]{{\includegraphics[width=0.35\linewidth]{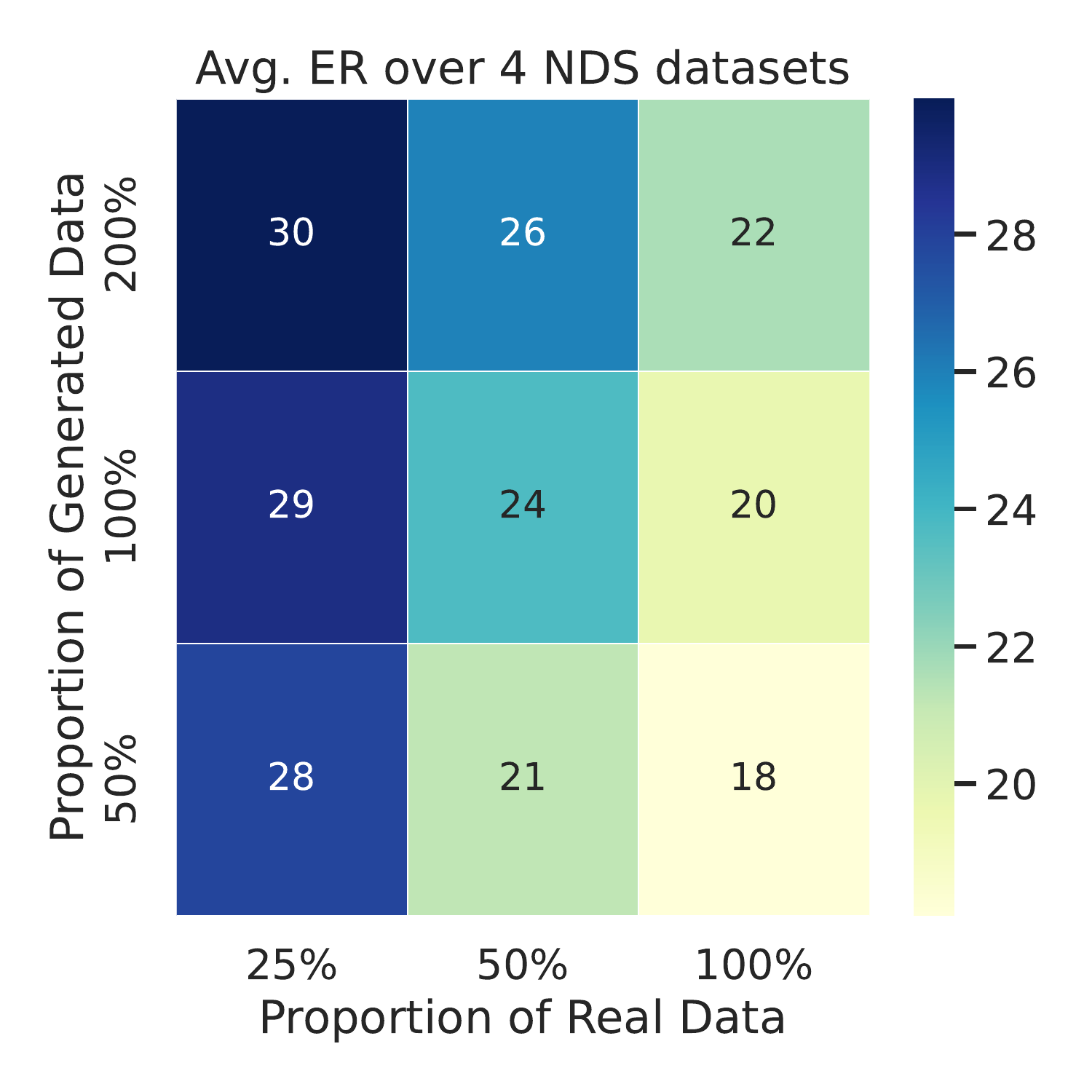}}}
    \caption{Variation in the accuracy and the effective robustness on ImageNet-R as we vary the proportion of the real ImageNet-100 data and the generated data created using its class labels in the training set. Here $100\%$ refers to 130K training size. While calculating effective robustness, standard training is performed on $100\%$ real data.}
    \label{appen_fig:variation_rendition}
\end{figure}

\begin{figure}[h]
    \centering
    \subfloat[\centering\label{appen_fig:av_acc_v2} Accuracy]{{\includegraphics[width=0.35\linewidth]{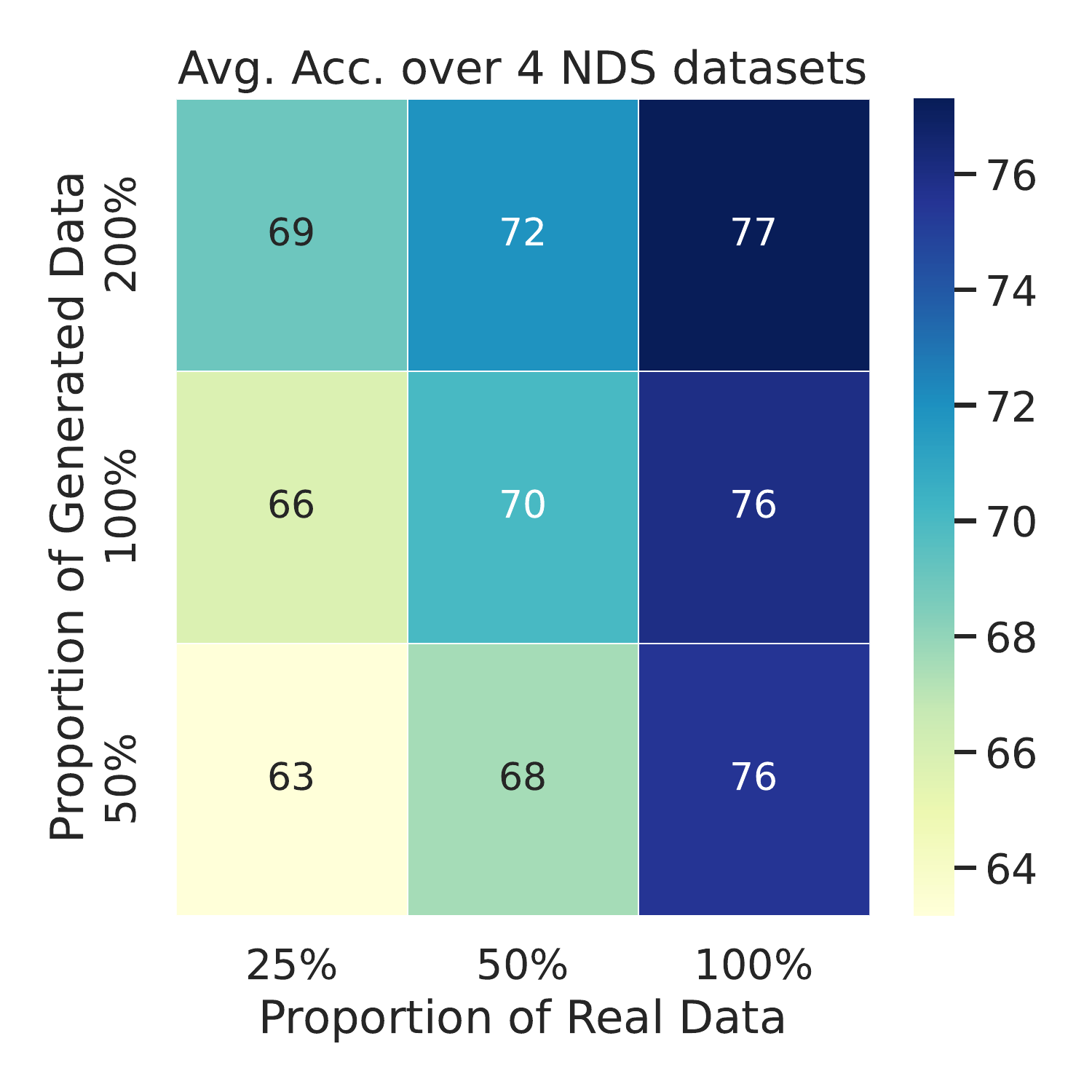}}}
    \subfloat[\centering\label{appen_fig:av_eff_v2}  Effective Robustness]{{\includegraphics[width=0.35\linewidth]{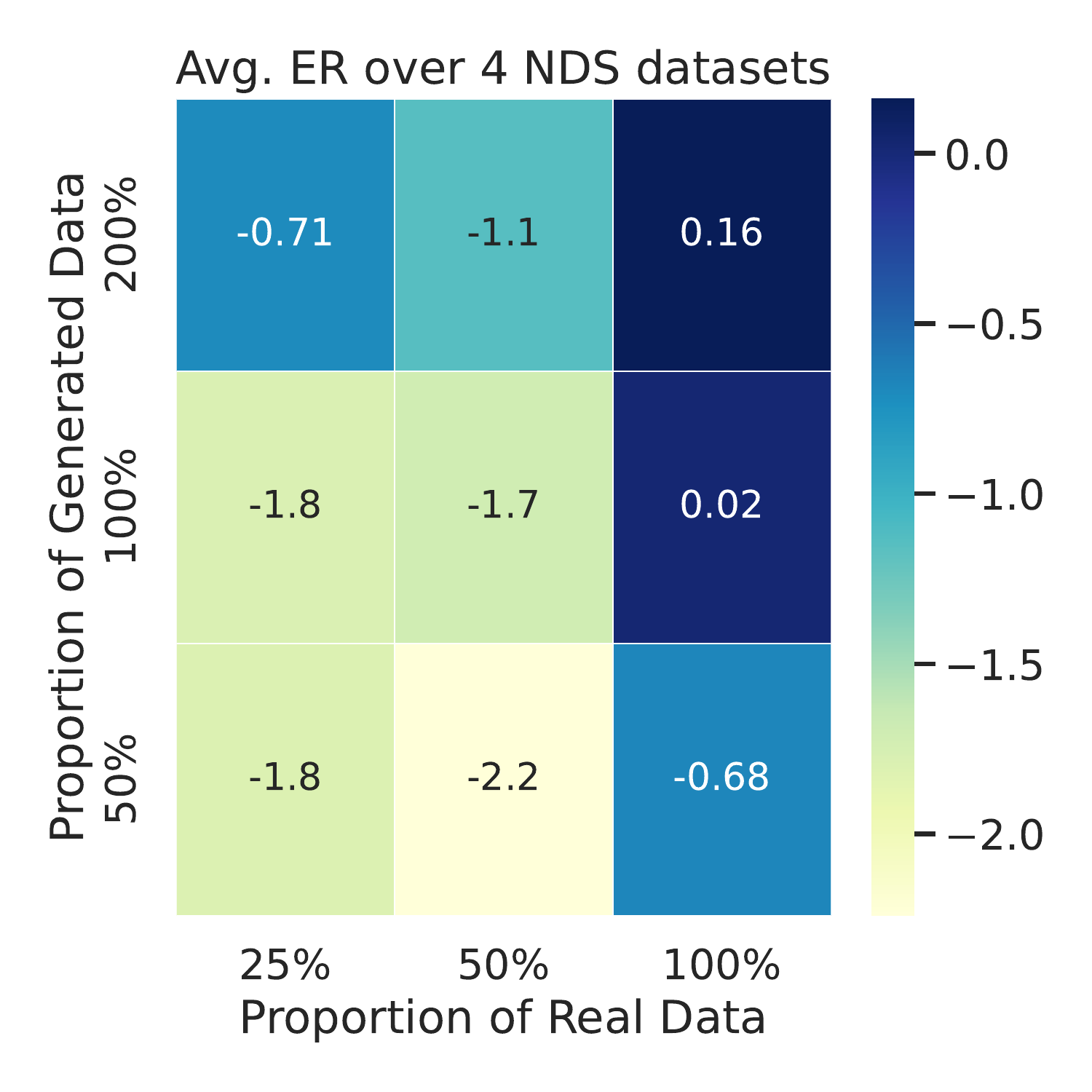}}}
    \caption{Variation in the accuracy and the effective robustness on ImageNet-V2 as we vary the proportion of the real ImageNet-100 data and the generated data created using its class labels in the training set. Here $100\%$ refers to 130K training size. While calculating effective robustness, standard training is performed on $100\%$ real data.}
    \label{appen_fig:variation_v2}
\end{figure}

\begin{figure}[h]
    \centering
    \subfloat[\centering\label{appen_fig:av_acc_} Accuracy]{{\includegraphics[width=0.35\linewidth]{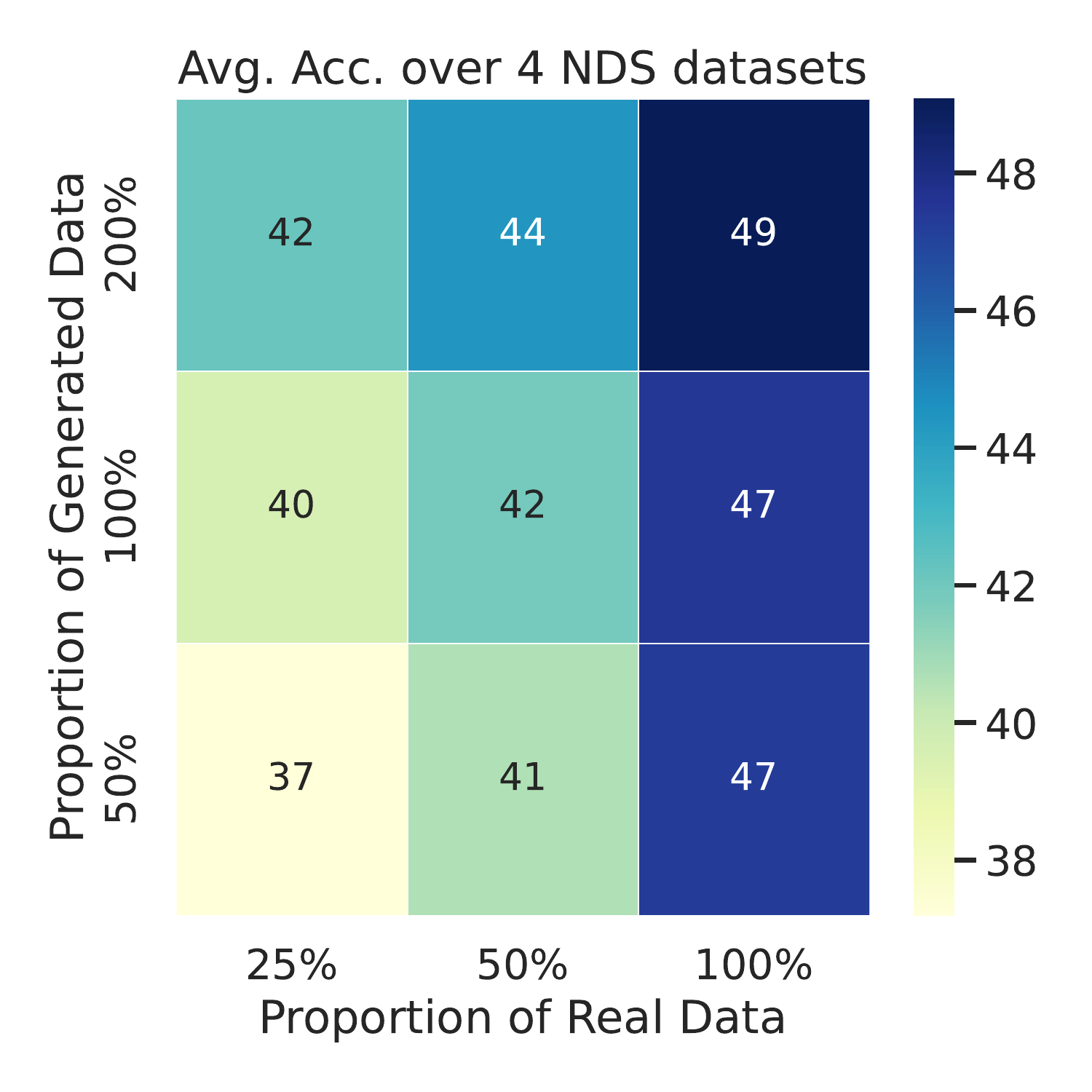}}}
    \subfloat[\centering\label{appen_fig:av_eff_s}  Effective Robustness]{{\includegraphics[width=0.35\linewidth]{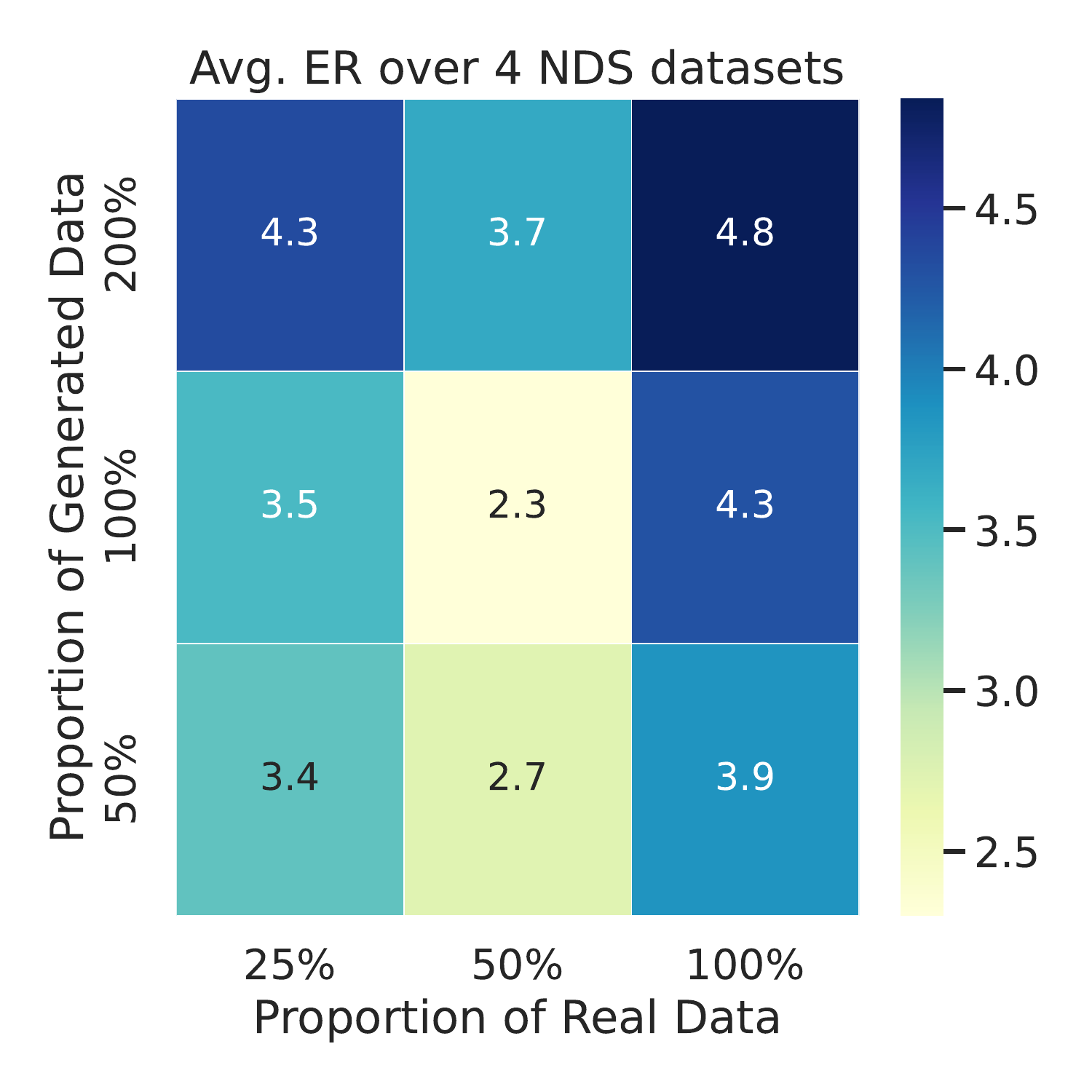}}}
    \caption{Variation in the accuracy and the effective robustness on ObjectNet as we vary the proportion of the real ImageNet-100 data and the generated data created using its class labels in the training set. Here $100\%$ refers to 130K training size. While calculating effective robustness, standard training is performed on $100\%$ real data.}
    \label{appen_fig:variation_objectnet}
\end{figure}

We present the effect of variation in the training size along the dimensions of the training data and the generated data in Figure \ref{appen_fig:variation_sketch}, \ref{appen_fig:variation_rendition}, \ref{appen_fig:variation_v2}, \ref{appen_fig:variation_objectnet}.

\subsection{Fixing the Amount of Training Data}

\begin{figure}[h]
    \centering    \subfloat[\centering\label{exp_fig:fixed_acc} Average Accuracy]{{\includegraphics[width=0.5\linewidth]{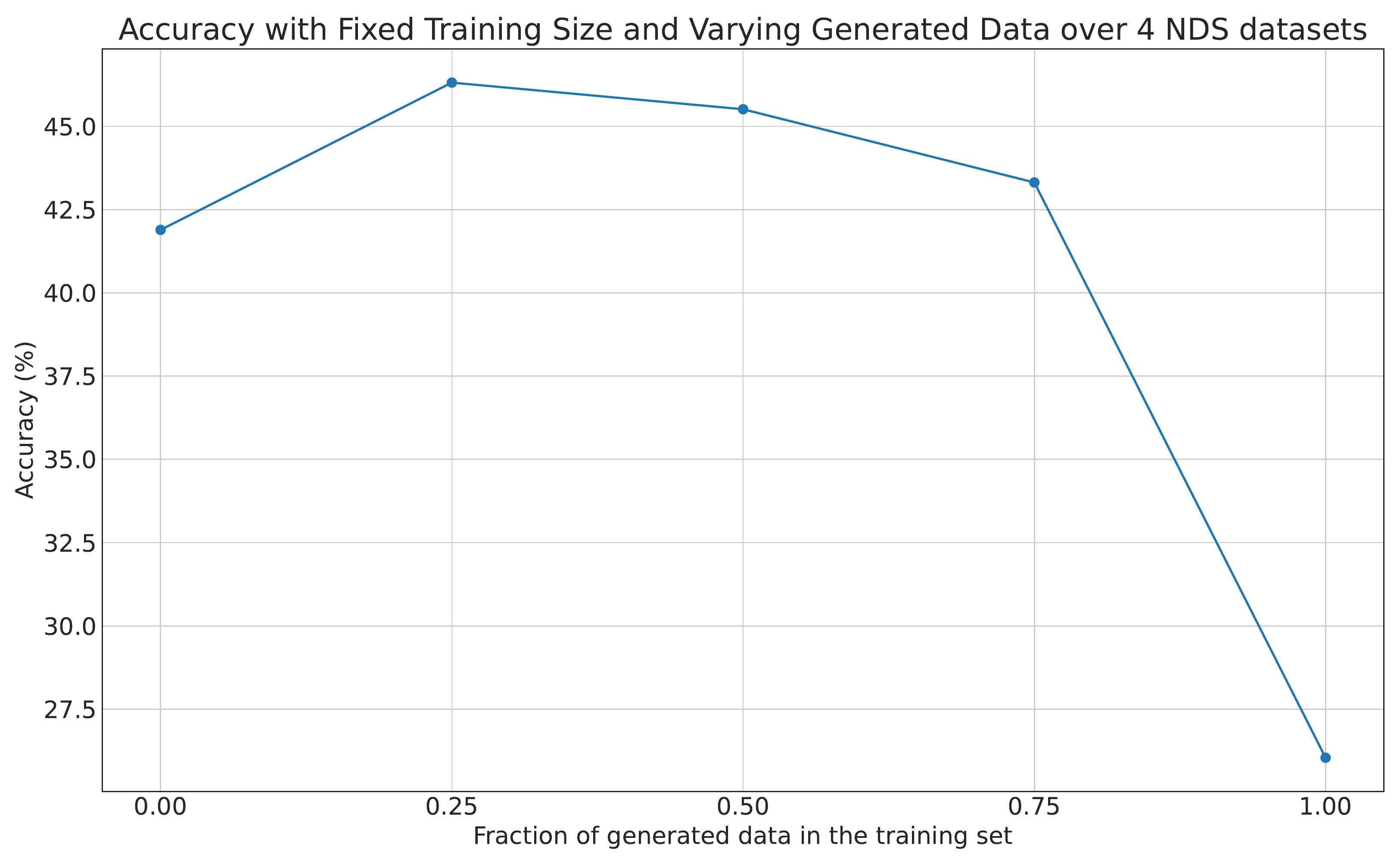}}}
    \subfloat[\centering\label{exp_fig:fixed_er} Average Effective Robustness]{{\includegraphics[width=0.5\linewidth]{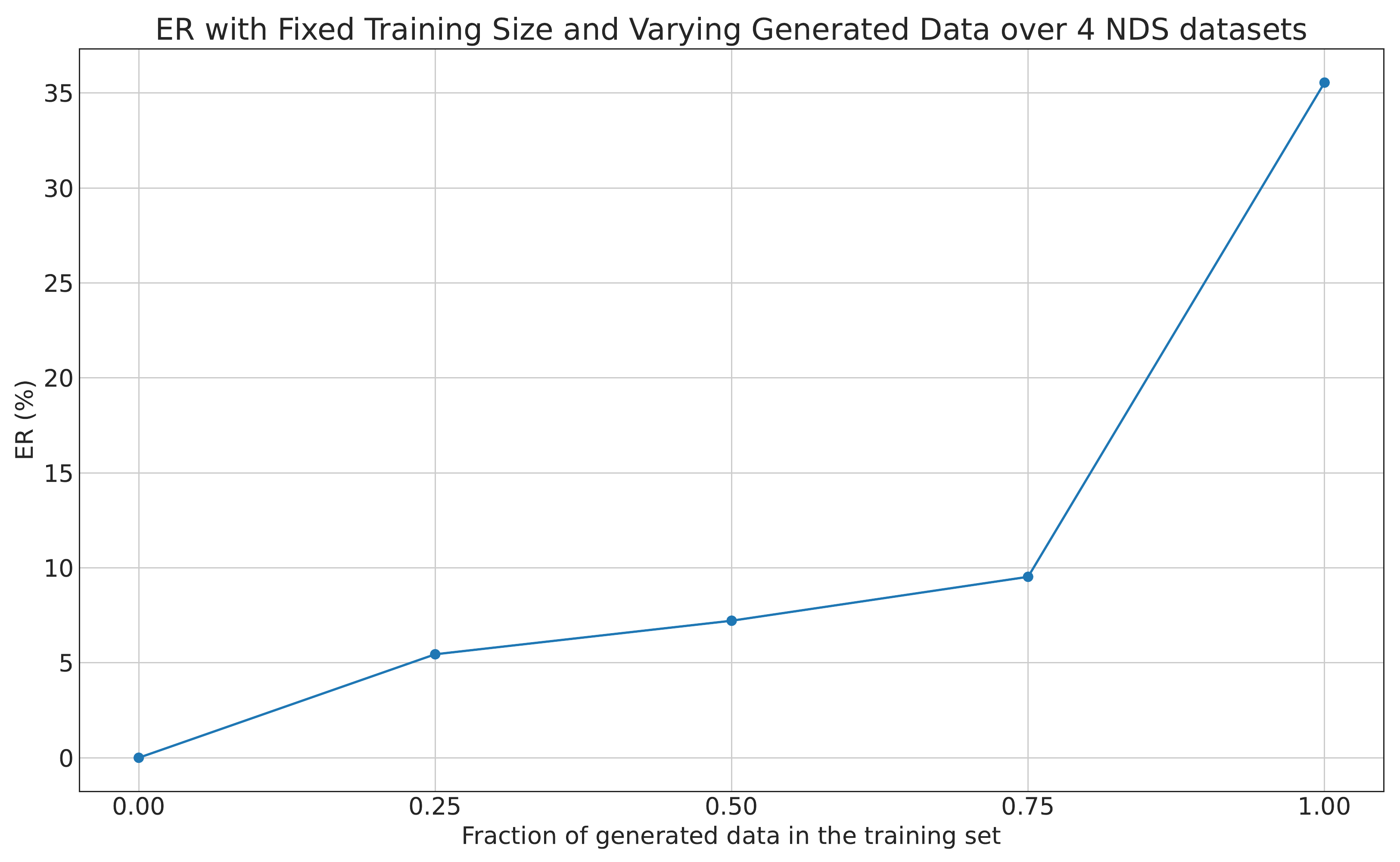}}}
    \caption{Variation in the accuracy and the effective robustness as we vary the proportion of the generated ImageNet1K data by fixing the number of training samples to 1.3M. While calculating effective robustness, standard training is performed on 1.3M real data. We report the results for the ResNeXt-50 classifier over three random seeds.}
    \label{appen_exp_fig:fixed_data}
\end{figure}

We conducted an experiment to examine the impact of varying the amount of generated data with a fixed 1.3M training sample budget on ImageNet1K. Figure \ref{appen_exp_fig:fixed_data} shows the accuracy and robustness of ResNeXt-50 averaged over four the natural distribution shift datasets. In Figure \ref{exp_fig:fixed_acc}, accuracy increases initially with increasing generated data but drops by $15\%$ when the fraction of generated data increases from 0.75 to 1. Conversely, in Figure \ref{exp_fig:fixed_er}, the effective robustness increases monotonically with the increase in the proportion of generated data in the training mixture.

% \begin{table}[h]
% \begin{center}
% \caption{Comparison of diversity for various generation strategies.}
% \begin{tabular}{l|c}
% \hline
%     \textbf{Data} & \textbf{Diversity}  \\\hline
%    Real    & 0.30 \\\hline
%    SD-Labels (\small{Diverse Templates}) & 0.26 \\
%    SD-Labels (\small{a photo of a \{class label\}}) & 0.15 \\
%    SD-Labels (\small{a rendition of a \{class label\}}) & 0.16 \\
%    SD-Images & \textbf{0.32}\\
%    SD-Labels and Images (\small{Diverse Templates}) & 0.23 \\
% \hline
% \end{tabular}%
% \label{exp_table:diversity_appendix}
% \end{center}
% \end{table}

\begin{figure}[h]
    \centering
    \includegraphics[width=0.5\linewidth]{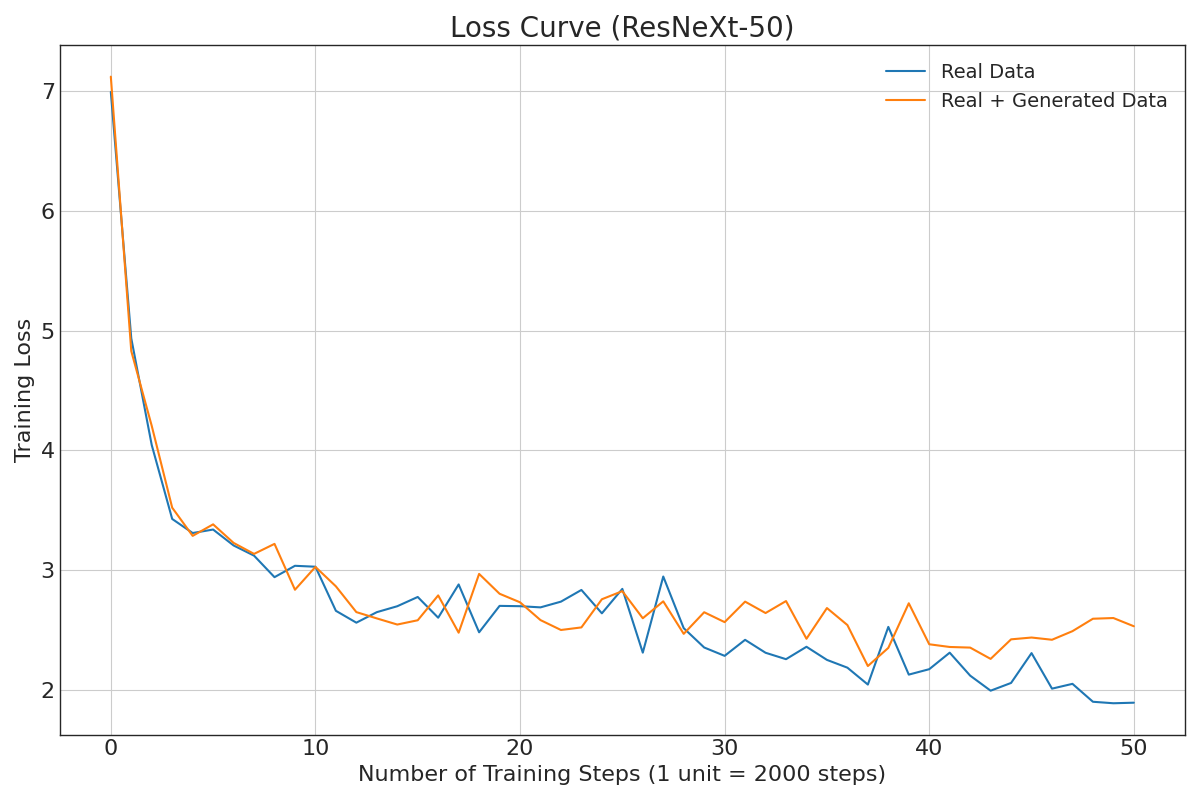}
    \caption{Comparison of the Loss Curve for ResNeXt-50 while training with the real and the generated data. The number of training samples in the real data is 1.3M whereas the number of training samples in the real and generated data scenario is 2.6M.}
    \label{appen_fig:training_curve}
\end{figure}

\section{Training Dynamics}
\label{app:training_dynamics}

We present the loss curve, in Figure \ref{appen_fig:training_curve}, to compare the training dynamics of a classifier, ResNeXt-50, on the real ImageNet-1K data and an equal mix of real and generated ImageNet-1K data in 100:100 proportion.

\end{document}